\documentclass[10pt,twocolumn]{article}
\usepackage{arxiv}

\usepackage[utf8]{inputenc} 
\usepackage[T1]{fontenc}    
\usepackage{hyperref}       
\usepackage{url}            
\usepackage{booktabs}       
\usepackage{amsfonts,amsmath}       
\usepackage{nicefrac}       
\usepackage{microtype}      
\usepackage{cleveref}       
\usepackage{lipsum}         
\usepackage{graphicx}
\usepackage{cite}
\usepackage{doi}
\usepackage{xcolor}
\usepackage{float}
\usepackage[font={sf,scriptsize},           
          labelfont={bf,color=black},
          caption=false]                  
         {subfig} 
\usepackage[font={it,small},labelfont=bf]{caption}
\usepackage{derivative} 
\usepackage{mathtools} 
\usepackage{dblfloatfix}

\newcommand{\myarrowparthalf}{\text{\raisebox{0.143cm}{\resizebox{0.5cm}{0.18cm}{--}}}}
\newcommand{\myarrowpart}{\myarrowparthalf\kern-0.41cm\raisebox{-0.169cm}{\kern-0.09cm\myarrowparthalf}\kern-0.14cm}
\newcommand{\myarrow}{$\overset{\kern-0.35em\text{\raisebox{0.1cm}{\normalsize DiscoGrad}}}{\myarrowpart\myarrowpart\myarrowpart\kern-0.25cm\Rightarrow}$}

\usepackage{listings}
\lstdefinestyle{dgstyle}{
  language=C++,
  basicstyle=\footnotesize\ttfamily,
  numbers=left,
  numbersep=5pt,
  numberstyle=\footnotesize\ttfamily,
  morekeywords={adouble,sdouble,aparams,DiscoGrad,DiscoGradFunc},
  escapechar=\$,
  rulecolor=\color{black},
  commentstyle=\color{gray},
}
\definecolor{codehighlight}{HTML}{fff1b0}

\newcommand{\prog}{\ensuremath{\mathcal{P}}}
\newcommand{\norm}{\ensuremath{\mathcal{N}}}
\newcommand{\expval}[2][]{\mathbb{E}_{#1}\!\left[#2\right]}
\newcommand{\cpp}{C\raisebox{0.01cm}{\texttt{++}}}
\newcommand{\traffic}[1][]{\if\relax\detokenize{#1}\relax{TRAFFIC}\else{TRAFFIC~#1x#1}\fi} 
\newcommand{\epidemics}{{EPIDEMICS}}
\newcommand{\hotel}{{HOTEL}}
\newcommand{\ac}{{AC}}
\NewDocumentCommand{\grad}{e{_^}}{%
\mathop{}\! 
\nabla
\IfValueT{#1}{_{\!#1}} 
\IfValueT{#2}{^{#2}}   
\mkern1mu 
}
\NewDocumentCommand{\smgrad}{e{_^}}{%
\mathop{}\! 
\widetilde{\nabla}
\IfValueT{#1}{_{\!#1}} 
\IfValueT{#2}{^{#2}}   
\mkern1mu 
}

\newcommand{\secref}[1]{Section~\ref{#1}} 
\renewcommand{\eqref}[1]{Eq.~(\ref{#1})} 

\title{Smoothing Methods for Automatic Differentiation Across Conditional Branches}

\date{}
\usepackage{authblk}

\newbox{\orcid}\sbox{\orcid}{\includegraphics[scale=0.06]{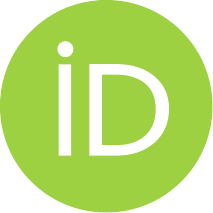}}
\author[1,2]{%
  \mbox{\href{https://orcid.org/0000-0002-4109-3608}{\usebox{\orcid}\hspace{1mm}Justin N.~Kreikemeyer}}%
}
\author[1,2]{%
  \mbox{\href{https://orcid.org/0000-0002-0211-7136}{\usebox{\orcid}\hspace{1mm}Philipp Andelfinger}}%
}
\affil[1]{%
  Institute for Visual and Analytic Computing, University of Rostock; 18059 Rostock, Germany.\hspace{2cm}
  \texttt{\{justin.kreikemeyer, philipp.andelfinger\}@uni-rostock.de}
}
\affil[2]{Both authors contributed equally to this work.}

\renewcommand{\undertitle}{Preprint}
\renewcommand{\headeright}{J.~N.~Kreikemeyer and P.~Andelfinger}
\renewcommand{\shorttitle}{Smoothing Methods for Automatic Differentiation Across Conditional Branches}

\hypersetup{
pdftitle={Smoothing Methods for Automatic Differentiation Across Conditional Branches},
pdfsubject={cs.LG,math.OC,eess.SY,cs.MS},
pdfauthor={Justin N.~Kreikemeyer, Philipp Andelfinger},
pdfkeywords={automatic differentiation, optimization, imperative programs, discontinuous control flow, parameter synthesis, gradient estimation, probabilistic execution, monte carlo approximation},
}

\begin{document}

\twocolumn[
\begin{@twocolumnfalse}

\maketitle

\vspace{-0.6cm}
\begin{abstract}
Programs involving discontinuities introduced by control flow constructs such as conditional branches pose challenges to mathematical optimization methods that assume a degree of smoothness in the objective function's response surface.
Smooth interpretation (SI) is a form of abstract interpretation that approximates the convolution of a program's output with a Gaussian kernel, thus smoothing its output in a principled manner.
Here, we combine SI with automatic differentiation (AD) to efficiently compute gradients of smoothed programs.
In contrast to AD across a regular program execution, these gradients also capture the effects of alternative control flow paths.
The combination of SI with AD enables the direct gradient-based parameter synthesis for branching programs, allowing for instance the calibration of simulation models or their combination with neural network models in machine learning pipelines.
We detail the effects of the approximations made for tractability in SI and propose a novel Monte Carlo estimator that avoids the underlying assumptions by estimating the smoothed programs' gradients through a combination of AD and sampling.
Using DiscoGrad, our tool for automatically translating simple \cpp\ programs to a smooth differentiable form, we perform an extensive evaluation.
We compare the combination of SI with AD and our Monte Carlo estimator to existing gradient-free and stochastic methods on four non-trivial and originally discontinuous problems ranging from classical simulation-based optimization to neural network-driven control.
While the optimization progress with the SI-based estimator depends on the complexity of the program's control flow, our Monte Carlo estimator is competitive in all problems, exhibiting the fastest convergence by a substantial margin in our highest-dimensional problem.
\end{abstract}


\vspace{0.8cm}

\end{@twocolumnfalse}
]

\section{Introduction}
\label{sec:introduction}

Parameter synthesis through optimization is a central task in fields such as modeling and simulation, control theory, and machine learning.
The difficulty of the optimization tasks increases with the number of parameters required to accurately model increasingly complex systems.
In the machine learning field, the well-known backpropagation algorithm~\cite{rumelhart1986learning} is commonly used for gradient-based training of deep neural network models across enormous numbers of parameters.
Gradient-based methods promise fast convergence to a local optimum, but require the existence and calculation of the optimization problem's partial derivatives.
Automatic differentiation (AD) techniques~\cite{griewank1996algorithm,margossian2019review} automatically calculate and propagate these derivatives through the arithmetic of arbitrary computer programs.

\begin{figure}[t!]
\vspace{0.65cm}
\includegraphics[width=0.99\columnwidth]{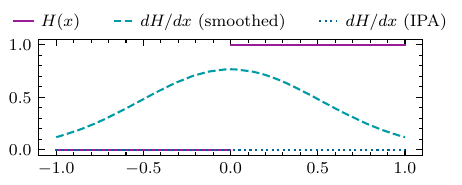}
\caption{Graph of the Heaviside step function and its derivative estimated pathwise (e.g., through automatic differentiation) by IPA, and by a smoothing estimator.}
\label{fig:ipa_fails}
\end{figure}

However, while these pathwise derivatives agree with the definition of the partial derivative, they do not provide sufficient gradient information if the control flow of the problem depends on the parameters.
As a characteristic example for branching control flow, consider the Heaviside step function $H(x)\,{=}\,\mathbf{1}_{x\geq 0}$, depicted in Fig.~\ref{fig:ipa_fails}.

This piecewise function has derivative $dH/dx\,{=}\,0$ everywhere, except at a discontinuity at $x\,{=}\,0$, where it is infinite, i.e., its derivative is the Dirac-delta.
Here, AD can correctly determine the derivative at any $x\,{\ne}\,0$, but its value of zero prohibits the use of gradient descent and does not provide any information about the jump discontinuity.
Evidently, this situation remains even if the derivative is averaged across different sample points in the parameter space (cf.~Fig.~\ref{fig:ipa_fails}, dotted line).
More generally, from the literature on infinitesimal perturbation analysis (IPA)~\cite{ho1983new} it is known that for stochastic discontinuous programs, simple averaging of the pathwise derivatives leads to a biased gradient estimator.
One solution is to calculate the gradient of a smooth approximation of $H$ (cf.~Fig.~\ref{fig:ipa_fails}, dashed line).

The need for differentiating discontinuous functions currently arises in many practical applications, such as neurosymbolic programming \cite{sun2022neurosymbolic}, program synthesis \cite{gaunt2016terpret}, agent-based simulation \cite{andelfinger2022towards}, and inverse rendering \cite{kato2020differentiable}. 
Thus, the challenge of obtaining gradients over discontinuities has been tackled from several angles: interpolation~\cite{andelfinger2022towards}, stochastic (Monte Carlo) estimation~\cite{suh2022bundled}, and smoothing over discrete randomness~\cite{arya2022automatic}.

Here, we explore two novel approaches for providing smoothed gradients of imperative programs involving branching control flow.
Using our tool DiscoGrad, problems involving \emph{parameter-dependent control flow} formulated in the \cpp\ language (cf.~Section~\ref{sec:discograd} for a brief description of supported language constructs) can be automatically differentiated to determine their smoothed gradients, enabling the use of gradient-based methods for local optimization.
As observed in the training of neural networks with backpropagation, suitable gradient descent algorithms are capable of finding high-quality solutions, even for non-convex functions \cite{choromanska2015loss}.
Accordingly, our methods assume neither smoothness nor convexity and we evaluate the performance of our proposed estimators on four high-dimensional, discontinuous optimization problems.
Our main contributions are:

\begin{enumerate}
  \item We show how the existing technique of Smooth Interpretation (SI) \cite{chaudhuri2010smooth}, a form of abstract interpretation, can be combined with AD to obtain gradients across discontinuities in \secref{sec:approach}.
  \item We provide a clear description of the assumptions made in SI's probabilistic execution of a program and their effects on the output's fidelity in \secref{sec:relaxing_sis_assumptions}.
  \item We propose a novel gradient estimator that avoids SI's assumptions by a combination of AD and Monte Carlo sampling in \secref{sec:monte_carlo_approach_to_smooth_differentiation}.
  \item We present DiscoGrad\footnote{Available at \url{https://doi.org/10.5281/zenodo.10288017}.}, our tool to automatically translate \cpp\ programs to an efficiently smoothed, differentiable counterpart using our proposed smoothing methods and other existing gradient estimators in \secref{sec:discograd}.
\item We provide an extensive evaluation of the estimator's execution times, gradient fidelity, and optimization progress against existing sampling-based schemes for local optimization such as REINFORCE~\cite{williams1992simple} and non-gradient based, global optimization methods (genetic algorithm, simulated annealing) in \secref{sec:evaluation}.
\end{enumerate}

In the following sections, we introduce AD, the smoothing of gradients, and SI (\secref{sec:background}) and review the related literature (\secref{sec:related_work}).
\secref{sec:smooth_automatic_differentiation} presents our main results.
Finally, we carry out an extensive evaluation (\secref{sec:evaluation}), concluding with final remarks and future directions (\secref{sec:conclusion}).

\section{Background}
\label{sec:background}

In the following, we outline the established work on differentiating programs, with a focus on programs involving branching control flow.
Starting from automatic differentiation as the base technique for differentiating programs, we introduce stochastic smoothing as well as smooth interpretation.

\subsection{Automatic Differentiation}
\label{sec:automatic_differentiation}

Automatic differentiation (AD) is a method to compute partial derivatives of computer programs~\cite{griewank1996algorithm,margossian2019review}.
Treating a program $\prog$ as a composition of mathematical functions $\prog\,{=}\,f_1\,{\circ}\,f_2\,{\circ}\cdots{\circ}\,f_n$, AD repeatedly applies the chain rule $f_i'\,{=}\,(f_{i+1}\circ f_{i+2})'\cdot f_{i+2}'$ to calculate the program's derivative $\prog'$ from the inputs $f_n$.
The well-known backpropagation algorithm~\cite{rumelhart1986learning} widely used in machine learning is a special case of AD.

The literature distinguishes two approaches: \emph{Forward-mode} AD propagates derivative information throughout the (forward) executions of the program by augmenting the involved variables $v$ with a so-called tangent value $\dot{v}$.
After each invocation of a mathematical function, this value is updated according to the function's derivative and the chain rule.
Thus, at any given point in the execution, the tangent value can be interpreted as the partial derivative $\partial f_j(\textbf{x})/\partial x_i$ of the operations up to the current point $j$ wrt.~the component $x_i$ of the input vector $\textbf{x}$.

In contrast, \emph{reverse-mode} AD records the arithmetic operations and values involved in the program's forward execution in a so-called \emph{tape} and computes the partial derivatives in a subsequent step by traversing the tape in reverse.
While forward-mode AD calculates the partial derivatives of a single input variable wrt.~all output variables throughout a single program execution, reverse-mode AD calculates the derivatives of all inputs wrt.~a single output based on a single traversal of the tape following the program's termination.
The pathwise gradients computed by AD are exact to machine precision.
This is in contrast to finite differences methods, whose fidelity depends on finding an appropriate step size.

Many programs encountered in optimization problems involve input-dependent control flow, which typically introduces jump discontinuities.
Unfortunately, as AD's purely arithmetic view of a single forward execution of a program cannot account for alternative control flow paths, it produces gradients of limited utility for optimization for such programs (cf.~Fig.~\ref{fig:ipa_fails}).
For stochastic programs, averaging across pathwise derivatives, known as infinitesimal perturbation analysis (IPA), produces an unbiased estimator only if the program permits the exchange of expectation and differentiation~\cite{gong1987smoothed}:
$$
\grad\expval{\prog(\mathbf{x})} = \expval{\grad\prog(\mathbf{x})}
$$
Programs involving input-dependent discontinuities typically violate this condition.
For deterministic programs, this form of smoothing can also be used by perturbing the input vector $\mathbf{x}$ with random noise.
Sampling-based estimators applicable to the discontinuous case are discussed in Sections~\ref{sec:sampling_based_gradient_estimation} and \ref{sec:gradient_estimators}.
As described next, the expected value can alternatively be obtained by a symbolic probabilistic execution.

\subsection{Smooth Interpretation} 

Smooth interpretation (SI)~\cite{chaudhuri2010smooth} is a method to smooth the output of programs involving discontinuities, making them more amenable to numerical optimization using black-box approaches such as Nelder and Mead's method~\cite{nelder1965simplex}.
Building on abstract interpretation~\cite{cousot1977abstract,smith2008probabilistic} and probabilistic program semantics~\cite{monniaux2000abstract}, SI executes a program $\prog$ according to a smoothed semantics that approximates the convolution of the program output with a Gaussian kernel $f_{\mathbf{x},\Sigma}$:
\begin{equation}\label{eq:smooth_program_si}
  \widetilde{\prog}(\mathbf{x})\coloneqq\int_{\mathbf{y}\in\mathbb{R}^n}\prog(\mathbf{y})f_{\mathbf{x},\Sigma}(\mathbf{y})d\mathbf{y}.
\end{equation}

\noindent Here, $\mathbf{x}$ is the program's $n$-dimensional input vector and $\Sigma$ a diagonal covariance matrix determining the amount of smoothing.

In SI, each originally scalar input variable $x_i\,{\in}\,\mathbf{x}$ is substituted by a Gaussian random variable $X_i$ with mean ${\mu_x}_i\,{=}\,x_i$ and a configurable standard deviation ${\sigma_x}_i$ sometimes referred to as the smoothing factor.
Now, the arithmetic operations specified in the program operate on and generate random variables.
As an approximation, the distribution of any operation's output is in practice again represented by a Gaussian characterized by its mean and standard deviation.
Thus, the output variables of a smooth interpretation are, just like the inputs, Gaussian random variables.
The result of the interpretation, i.e., of the approximate convolution of the Gaussian with the program at the current input, are the expectations of the output variables.

A key aspect of SI is the handling of conditional branches, as encountered in the form of \texttt{if-else} statements.
The smoothed semantics require both possible paths to be executed and weighted according to the distribution of the variables involved in the branching condition.
This leads to two possible distributions for some of the variables.
Hence, each variable is represented as a mixture distribution, each element of which represents a Gaussian approximation of the distribution resulting from one of the program's (sequences of) branches.
To limit the number of elements of each mixture distribution, a ``Restrict'' algorithm combines the results of branches in a way that minimizes the deviation from the original overall mixture distributions of the variables.

In general, the exact convolution of a program with a Gaussian is intractable.
The approximations made by SI, e.g., assuming the program state to be a Gaussian mixture and limiting it to a finite size through Restrict, enable a practical application of the method and will be further explored in \secref{sec:relaxing_sis_assumptions}.

\section{Related Work}
\label{sec:related_work}

Rooted in the field of non-smooth optimization~\cite{gaudioso2022essentials}, the (gradient-based) optimization of discontinuous programs has recently seen major interest across many domains, for example, machine learning~\cite{petersen2022learning}, computer graphics~\cite{kato2020differentiable} and optimal control \cite{suh2022bundled}.
Besides gradient-free approaches such as genetic algorithms or the Nelder-Mead method~\cite{nelder1965simplex}, the state of the art in non-smooth optimization includes bundle methods, which augment the subgradient method through the exploitation of past subgradient information \cite{makela2002survey} and gradient sampling methods exploiting piecewise differentiability~\cite{burke2020gradient}.
In contrast, we consider smoothed gradients of problems that, while piecewise differentiable, typically provide only zero-vector gradients (cf.~Fig.~\ref{fig:ipa_fails}).
For these, pathwise estimates based on pathwise gradients, as in IPA~\cite{ho1983new}, are insufficient.

In the following, we discuss existing work on differentiating \emph{across} branching control flow directly related to ours.
Broader overviews of gradient estimation techniques are given in~\cite{fu2006chapter} and \cite{lecuyer1991overview}.

\subsection{Sampling-Based Gradient Estimation}
\label{sec:sampling_based_gradient_estimation}

Based on the conditional Monte Carlo method for variance reduction, \emph{smoothed perturbation analysis} (SPA) obtains an unbiased gradient estimate through conditional expectations~\cite{gong1987smoothed}.
By choosing suitable problem variables (called characterization) to condition on, the calculation of the expected value is effectively separated into continuous parts, allowing for the interchange of the expectation and differentiation operations (cf.\ the end of \ref{sec:automatic_differentiation}).
While SPA has been widely applied to differentiate discontinuous problems such as certain discrete-event simulations, its applicability is limited by the need to manually determine a suitable characterization to condition on for the problem at hand.
For an overview of the many variations of (S)PA refer to~\cite{fu2006chapter} (Section 9) and the references therein.

The \emph{REINFORCE} estimator, commonly employed in reinforcement learning, exploits the differentiation rule of the logarithm to eliminate the need for calculating gradients of the program~\cite{williams1992simple}.
The gradient is calculated from the plain program output, multiplied by the log-derivatives of the program-specific probability density (cf.~\secref{sec:gradient_estimators}).
REINFORCE is thus also referred to as the log-derivative trick, likelihood ratio estimator or score function estimator.
Similar to the characterization conditioned on in SPA, the probability density and its log-derivative are problem-specific.

Some problem-independent (black-box) estimators are given in~\cite{nesterov2017random}, therein referred to as \emph{gradient-free oracles}.
The authors analyze the convergence of a scheme introduced in Chapter 3.4 of~\cite{polyak1987introduction}, which estimates the descent direction through directional derivatives, calculated by a randomized version of finite differences where each input dimension is perturbed simultaneously.
We will make use of the first of these gradient-free oracles for comparison in the evaluation, where it is also briefly introduced in \secref{sec:gradient_estimators}.
Due to the lack of a commonly used name for this estimator, we abbreviate it as PGO (Polyak's Gradient-Free Oracle).
A similar construction using non-directional finite differences is proposed in~\cite{suh2022bundled}.

We note that sampling-based schemes for stochastic programs, like SPA and REINFORCE, can also be applied to the case of deterministic objective functions by introducing artificial perturbations to the program inputs.
If these perturbations are sampled from a normal distribution, their estimates approach the gradient of the convolution integral from \eqref{eq:smooth_program_si} as the number of samples approaches infinity.

\subsection{Combination of Sampling and AD}

Some recent works propose combinations of sampling-based methods with AD rather than finite differences.

In~\cite{arya2022automatic}, an unbiased SPA estimator was derived for programs involving discrete randomness and integrated with the AD process in the Julia package \texttt{StochasticAD}.
This allows for the automatic smooth differentiation of programs that sample from discrete probability distributions with parameters depending on the program inputs.
In contrast to our work, this approach does not consider input-dependent discrete control flow.

A recent use of forward-mode AD is found in the ``forward gradient'', which is determined by sampling over directional derivatives~\cite{baydin2022gradients}.
While their work does not consider differentiation across discontinuities, our methods share the use of forward-mode AD for problems where the reverse mode would be the conventional choice.
As we will see in \secref{sec:execution_time}, forward-mode AD incurs only a tolerable overhead in our benchmark problems, while allowing us to efficiently obtain intermediate partial derivatives and avoiding reverse-mode AD's linear dependence of the memory consumption on the program length.

Finally, a sampling-based method proposed in a recent preprint~\cite{christodoulou2023differentiable} achieves differentiability by applying a static degree of smoothing at each branch and systemically visiting all control flow paths whose probability is within machine precision.
Their approach shares with SI the challenge of scaling to problems with non-trivial numbers of branches without introducing biases, the effects of which on SI's fidelity are detailed in Section~\ref{sec:relaxing_sis_assumptions} and quantified in Section~\ref{sec:evaluation}.

\subsection{Differentiable Programming Languages and Neurosymbolic Programming}

Differentiable programming languages offer semantics that allow for a sound calculation of gradients across entire, typically functional, programs.
Abadi presented operational and denotational semantics for a functional language that includes a construct for reverse-mode AD~\cite{abadi2019simple}.
Discontinuities are ruled out by assuming that constructs such as conditional branches are substituted by smooth approximations by the user.

Some recent languages treat discontinuities natively.
Sherman et al.~presented semantics for a functional language that covers non-differentiable functions, but requires continuity~\cite{sherman2021computable}.
The functional language ADEV~\cite{lew2023adev}, which targets differentiable probabilistic programming, allows discontinuities to only depend on the program's stochasticity, not on the parameters.
This is the same condition that is satisfied after applying the reparametrization trick~\cite{kingma2013auto}.
A more general approach for handling discontinuities is taken in the functional language by Amorim et al.~\cite{amorim2022distribution}, which relies on distribution theory to soundly express the contribution of discontinuities to the gradient.
The resulting integrals are approximated by Monte Carlo sampling.
In contrast to these works focused on language semantics, we propose and study concrete gradient estimators that approximate gradients of smoothed imperative programs, specifically targeting the case where discontinuities depend on the parameters.

Among the use cases of differentiable programming languages is the gradient-based \emph{synthesis of symbolic programs}, which offers an alternative to traditional combinatorial program search.
The search over symbolic programs is achieved by interpreters that employ continuous relaxations to enable the computation of gradients of the program's output wrt.~its parameters, which may represent numerical constants, instructions, or the registers to operate on~\cite{feser2016differentiable,bovsnjak2017programming,gaunt2016terpret,gaunt2017differentiable}.
\emph{Neurosymbolic programming}~\cite{chaudhuri2021neurosymbolic,sun2022neurosymbolic} extends this idea towards programs combining symbolic and neural building blocks.
In a recent work, a generalization of the REINFORCE estimator was used to differentiate across symbolic program executions in order to determine parameters leading to control flow paths that adhere to a safety criterion~\cite{yang2022safe}.
Although the focus of our present work is on the parameter synthesis for existing programs, the proposed gradient estimators may benefit synthesis approaches as well.

\subsection{Domain-Specific Approaches}

Methods for gradient smoothing are proposed and applied in many contexts, motivated by a myriad of goals.
Here, we discuss relevant publications from the popular fields of neurosymbolic programming, program synthesis, differentiable rendering, and simulation-based optimization.

Some recent work achieves smoothing by weighted averaging of variable values across branches as a basis for combining neural networks with traditional algorithms~\cite{DBLP:journals/corr/abs-2007-12101,petersen2019algonet}, for parameter estimation across agent-based simulations~\cite{andelfinger2022towards}, and for antialiasing~\cite{yang2018approximate}.
In contrast to SI and unbiased black-box estimators, these works lack a well-defined probabilistic semantics and thus do not offer a clear interpretation of the smoothed output.

An alternative approach common in simulation-based optimization is to sample a model's input-output relation to generate a \emph{surrogate model}~\cite{barton2020tutorial}.
Depending on the type of surrogate, e.g., a neural network, the resulting model may be smooth and differentiable.
A similar approach has been taken in \emph{systems security} for gradient-based fuzzing~\cite{she2019neuzz}.
Surrogate models are typically fitted to input-output samples in a black-box fashion, after which gradient estimates are made without involvement of the original model.
In contrast, the gradient estimators proposed in our work operate on the original program, making use of its internal structure.

Finally, the field of computer graphics has also shown broad interest in the differentiation of discontinuous programs.
Differentiable rendering~\cite{kato2020differentiable} aims at determining partial derivatives of pixel values with respect to scene parameters, enabling applications such as inverse rendering, i.e., determining scene parameters that best fit real-world image data.
Modern rendering techniques are typically based on Monte Carlo sampling of light rays through three-dimensional scenes.
The integrals approximated in this manner may carry discontinuities related to the visibility of objects in the scene.
This problem can be solved either by explicitly sampling edges that cause the discontinuities~\cite{li2018differentiable}, or more scalably by applying problem-specific reparametrizations to the objective function so that the position of discontinuities becomes independent of the parameters~\cite{loubet2019reparameterizing}.
An overview of Monte Carlo techniques for differentiable rendering is given by Zeltner et al.~\cite{zeltner2021monte}.

In contrast to the above works, the gradient estimators proposed and evaluated in the remainder of our article target generic imperative programs without reliance on domain-specific problem properties.

\section{Smooth Automatic Differentiation}
\label{sec:smooth_automatic_differentiation}

Many optimization problems are naturally formulated as imperative programs.
However, the existing work on smooth differentiation lacks a method that 1.~focuses on imperative programs involving conditional branching on the input values and 2.~makes use of exact pathwise derivatives as determined via AD.
In this central section, we will show two possible candidates.
First, we provide a framework for the problem of calculating the smoothed gradient of discontinuous programs where the conditional control flow depends on the input vector.
By mapping SI to this framework, its integration with AD becomes straight forward, providing our first (deterministic) estimator.
However, this approach is encumbered by the strong assumptions of SI.
We explore ways to relax these through information gained by AD, but find that the benefits of these improvements are dwarfed by the effect of SI's restricted representation of the probabilistic program states.
Thus, our second (sampling-based) candidate is an AD-powered Monte Carlo approach operating under much lighter assumptions, often yielding significantly more accurate gradient estimates.

\subsection{Approach}
\label{sec:approach}

In our derivations, we consider optimization problems expressed as imperative programs $\prog{:}\ \mathbb{R}^n\,{\rightarrow}\,\mathbb{R}$ mapping $n$ input variables to a single output value.
Thus, \prog\ is typically a piecewise function.
We further assume that discontinuities only arise through branch and loop conditions, noting that common discontinuous functions such as the absolute value function can be rewritten using conditional branching.

As shown in \cite{chaudhuri2010smooth}, the smoothing of \prog\ with a (multivariate) Gaussian kernel $f_{\mathbf{x},\Sigma}$ can be expressed in terms of a convolution
\begin{equation}\label{eq:smooth_program_as_convolution}
  \widetilde{\prog}(\mathbf{x})\coloneqq\int_{\mathbf{y}\in\mathbb{R}^n}\prog(\mathbf{y})f_{\mathbf{x},\Sigma}(\mathbf{y})d\mathbf{y} = \expval[X\sim\mathcal{N}(\mathbf{x},\Sigma)]{\prog(X)},
\end{equation}

\noindent where $\mathbf{x}$ is the program's $n$-dimensional input vector and $\Sigma$ a diagonal covariance matrix determining the amount of smoothing.
This form of smoothing is sometimes also called the (generalized) Weierstrass transform.
By the law of the unconscious statistician, the convolution \eqref{eq:smooth_program_as_convolution} can be regarded as taking the expected value of the program's output distribution when executed on normally distributed random variables $X\,{\sim}\,\norm(\mathbf{x},\Sigma)$, see also \cite{fu2006chapter}, Section 4.
It is important to note that in our case randomness is introduced by moving from $\textbf{x}$ to $X$, i.e., our derivations are also applicable to deterministic programs.

A possible strategy to automatically calculate the smoothed gradient $\widetilde{\grad}\prog(\textbf{x})\,{\coloneqq}\,\grad\widetilde{\prog}(\textbf{x})\,{=}\,\grad_{\mathbf{x}}\expval{\prog(X)}$ is to exchange the expectation and gradient operators, as is done in IPA.
While this enables a close and automatic approximation, e.g., through a Monte Carlo approach and AD, the equality $\grad_{\mathbf{x}}\expval{\prog(X)}\,{=}\,\expval{\grad_{\mathbf{x}}\prog(X)}$ only holds if $\prog$ is continuous (precise conditions are given in \cite{lecuyer1990unified}).

In the case of imperative programs with control flow depending on $\textbf{x}$, $\prog$ is not generally continuous, requiring an alternative approach.
We observe that the control flow partitions the program's (discontinuous) output $\prog(\textbf{x})$ into several continuous parts.
In the probabilistic execution context, we can isolate these continuous parts by conditioning the distribution of $\prog(X)$ on the execution path $p\,{\in}\,\{1,\dots,N\}$.
More precisely, let the random variable $\mathfrak{P}$ reflect which control flow path $p$ was taken in the execution of $\prog$.
Then, using the law of total expectation, we can decompose the integral \eqref{eq:smooth_program_as_convolution} into a sum over expectations of its path-specific outputs:
\begin{equation}\label{eq:discrete_smooth_program}
\begin{split}
  \widetilde{\prog}(\textbf{x}) & = \expval[X\sim\mathcal{N}(\mathbf{x},\Sigma)]{\prog(X)} \\
                                & = \mathbb{E}_\mathfrak{P}\!\bigl[\expval[X]{\prog(X)|\mathfrak{P}}\bigr]\\
                                & = \sum_{p=1}^N \mathbb{P}(\mathfrak{P}=p) \expval[X]{\prog(X)|\mathfrak{P}=p}.
\end{split}
\end{equation}

\noindent In practice, $\mathfrak{P}$ is defined in terms of the conjunction of branching conditions on $X$ encountered along each control flow path $p$.
We note that the idea of using conditional expectations to obtain a smooth objective is very similar to SPA (cf.\ the overview in \cite{lecuyer1991overview}).
Here, however, the conditioning does not justify the application of IPA, but rather provides a descriptive transformation.
\noindent Considering \eqref{eq:discrete_smooth_program}, the smoothed gradient of the program is given by:
\begin{equation}\label{eq:discrete_smooth_gradient}
\begin{split}
  \smgrad_{\mathbf{x}} \prog(X) & = \grad_{\mathbf{x}} \mathbb{E}_\mathfrak{P}\!\bigl[\expval[X\sim\mathcal{N}(\mathbf{x},\Sigma)]{\prog(X)|\mathfrak{P}}\bigr] \\
                                & = \grad_{\mathbf{x}}\sum_{p=1}^N \mathbb{P}(\mathfrak{P}=p) \expval[X]{\prog(X)|\mathfrak{P}=p} \\
                                & = \sum_{p=1}^N \big( \grad_{\mathbf{x}}\mathbb{P}(\mathfrak{P}=p) \big) \expval[X]{\prog(X)|\mathfrak{P}=p} + \\
                                & \phantom{=\,\,} \sum_{p=1}^N \mathbb{P}(\mathfrak{P}=p) \big( \grad_{\mathbf{x}}\expval[X]{\prog(X)|\mathfrak{P}=p} \big).
\end{split}
\end{equation}

\begin{figure*}[t]
  \centering
  \includegraphics[width=0.9\textwidth]{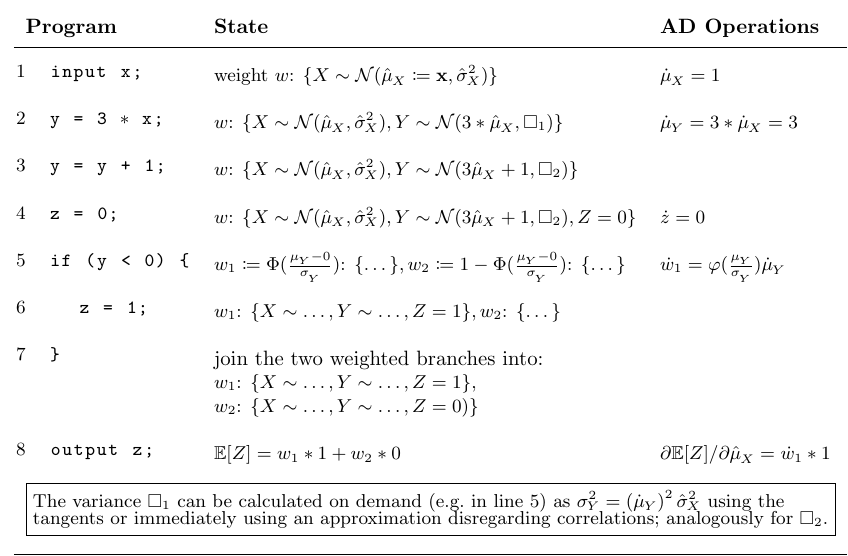}
  \caption{Example program (left) execution showcasing the probabilistic semantics of SI (center) and their integration with forward-mode AD (right). Only relevant AD operations are shown. The tangents $\dot v$ denote the (generally partial) derivative $\partial v/\partial \hat{\mu}_X^{}$ wrt.\ the mean $\mu$ of the normally distributed random variable ${X}\,{\sim}\,\mathcal{N}(\hat{\mu}_X^{},\hat{\sigma}_X^2)$. The hat $\hat{\ }$ symbol indicates an input value. Upon initialization, the (here scalar) input vector $\mathbf{x}$ is taken as the mean $\hat{\mu}_X^{}$; $\varphi$ and $\Phi$ denote the normal distribution's probability and cumulative density functions respectively. Note that in this example the pathwise derivative is 0, but through the combination of SI and AD, the derivative wrt.\ the branching condition is obtained.}
  \label{fig:the_figure}
\end{figure*}

\noindent Taking into account the sum and chain rules of differentiation, this requires determining the gradients of $\mathbb{P}(\mathfrak{P}\,{=}\,p)$ and $\expval{\prog(X)|\mathfrak{P}=p}$ wrt.\ $\mathbf{x}$.
In the general case, where the former does not depend on $\textbf{x}$, it is always $\textbf{0}$ and can be omitted.
However, in our case the probability of taking a branch is influenced by the branching condition, which depends on $\textbf{x}$, yielding a non-zero gradient vector.
The second problem with estimating \eqref{eq:discrete_smooth_gradient} is the number of possible paths $N$, which grows exponentially with the number of branches, leading to an exponential explosion of summation terms.

SI handles both problems by assuming that $\prog(X)$ follows a Gaussian mixture distribution of maximum size $M \ll N$.
The smoothed execution then involves propagating the first two moments $\textbf{x}$ and $\Sigma$ of $X$ and descendant variables through the program.
Fig.~\ref{fig:the_figure} (left and center) showcases this on a simple example program.
At branches, the mixture then naturally arises (for each variable) from the two new possible distributions of the \texttt{then} and \texttt{else} cases.
The weights of the two new mixture elements are calculated by evaluating the cumulative normal density parametrized with the two known moments of the input distribution.
To ensure that at most $M$ control paths are carried along, selected mixture elements are merged (cf.~\secref{sec:state_restriction_strategies}).
Thus, both the calculations leading to $\mathbb{P}(\mathfrak{P}\,{=}\,p)$ and $\expval{\prog(X)|\mathfrak{P}\,{=}\,p}$, which coincide with the weight and mean of the mixture element corresponding to $p$, are smooth functions of the program input.
We make use of this property by differentiating SI's approximations of the per-path weight and output via AD.
The right-hand side of Figure~\ref{fig:the_figure} shows the differentiation by the example of forward-mode AD, which tracks the derivatives wrt.\ the first moments $\mu_X^{\,}$ of $X$.
We now briefly consider some interesting opportunities opened up by this integration of AD.

\subsection{Relaxing SI's Assumptions}
\label{sec:relaxing_sis_assumptions}

The method of SI~\cite{chaudhuri2010smooth} imposes several major assumptions upon programs, trading off fidelity for execution speed:
\begin{enumerate}
  \item \emph{No interdependencies}: The inputs to any operation are assumed to be independent normal distributions.\label{enum:si_assumption_1}
  \item \emph{Everything is Gaussian}: The output of any operation is assumed to follow a normal distribution, which is not true in the general case, even assuming \ref{enum:si_assumption_1}).
    As a consequence, within a particular branch, the distribution of a variable $X_i$ (and its dependent variables) contains values excluded by the branching condition $X_i\leq c$.
    \label{enum:si_assumption_2}
  \item \emph{No truncation at branches}: When splitting the state at a branching point, the resulting two distributions are approximated by scaling the weight of the Gaussian mixture elements.
  Thus, their means and standard deviations remain unaltered, where in reality the results are the lower and upper tails of Gaussian distributions, i.e., truncated Gaussians.\label{enum:si_assumption_3}
\item \emph{Fixed-size state representation}: The program state across all possible control flow paths, whose number is exponential in the number of branches, is approximated by a fixed-size Gaussian mixture.\label{enum:si_assumption_4}
\end{enumerate}

\noindent While these assumptions permit a reasonable approximation of small programs with mostly affine operations, they cause substantial deviations for larger programs.
When integrating AD as described above, the resulting gradient estimates can thus become inaccurate and noisy (cf.~\secref{sec:evaluation}).
In the following, we briefly sketch how the additional information available via AD makes it possible to relax assumptions \ref{enum:si_assumption_1} to \ref{enum:si_assumption_3}, but show that the effects of \ref{enum:si_assumption_4} dominate the error.

Improvements regarding assumptions~\ref{enum:si_assumption_1} and \ref{enum:si_assumption_2} concern operations of the form $C\,{=}\,h(A,B)$.
We restrict this example to binary operations, as the n-ary case is analogous.
As shown in Figure~\ref{fig:the_figure}, SI determines the mean $\mu_C$ and variance $\sigma^2_C$ of $C$ from $A\,{\sim}\,\norm(\mu_A,\sigma^2_A)$ and $B\,{\sim}\,\norm(\mu_B,\sigma^2_B)$:
\begin{equation}
\begin{split}
  \mu_C      & = h(\mu_A, \mu_B) \\
  \sigma^2_C & = \left(\pdv{\mu_C}{\mu_A}\right)^{\!2}\sigma^2_A + \left(\pdv{\mu_C}{\mu_B}\right)^{\!2}\sigma^2_B
\end{split}
\end{equation}

\noindent This is a standard result of uncertainty propagation (UP)~\cite{benke2018error} and exact if $h$ is an affine function and $A$ and $B$ are independent.
Linear dependencies (correlations) between $A$ and $B$ can be accounted for by using the information obtained by AD (cf.\ also the explanation in Fig.~\ref{fig:the_figure}):
\begin{equation}\label{eq:uncertainty_propagation}
  \sigma^2_C = \sum_{i=1}^n\left(\pdv{\mu_C}{\mu_{X_i}^{\,}}\right)^{\!2}\sigma^2_{X_i},
\end{equation}

\noindent where $n$ is the number of inputs and the partial derivatives are provided by AD (see Appendix~\ref{app:implicit_covariance_through_ad} for a derivation).
Intuitively, this calculates the variance of $C$ from the variance of the inputs $X$, based on the transformations captured by the gradient since the start of the program and leading to $g$, which implicitly accounts for the covariance between $A$ and $B$.
Closer and automatically differentiable approximations of the distributions resulting from non-affine operations could be determined using higher-order Taylor approximations~\cite{yang2018approximate}, at the cost of additional computational overhead.

Lifting Assumption \ref{enum:si_assumption_2} poses the largest challenge, as the assumption of Gaussian distributions for all variables is the main enabler of an easy integration with AD.
To exclude invalid intervals from the variables' distributions on each path state, interval bounds could be carried along with the mean and variance for each mixture element~\cite{yang2022safe}. 
Moving beyond Gaussian distributions would require the tracking and updating of higher-order moments or explicit representations of the distributions' shapes.

Assumption~\ref{enum:si_assumption_3} can be lifted by calculating the truncated Gaussian distributions resulting from a branch and approximating them by their first two moments.
This requires obtaining the dependence of every variable on the branching condition to determine the point of truncation for each variable, which may be approximated by first-order dependencies determined via AD.

\begin{figure}[t!]
\includegraphics[width=0.99\columnwidth]{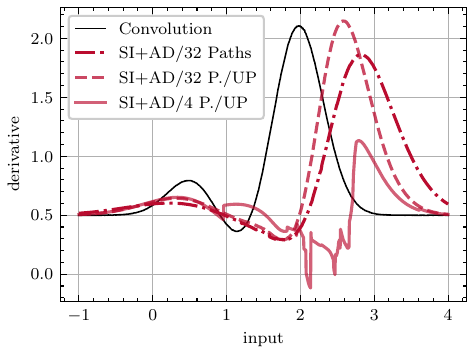}
\caption{Comparison of gradient fidelity wrt.\ the convolution with the original SI proposal with 32 tracked control flow paths and with correlation-preserving variance calculation using AD (uncertainty propagation, UP). When reducing the number of tracked paths to a small subset, as would be required for larger programs, the assumption of a fixed size mixture dominates the error. The non-smooth merging of mixture elements causes the gradient to jump or even assume the wrong sign, which is problematic for gradient descent.}
\label{fig:si_assumptions_test}
\end{figure}

To explore the effects of the enhancements on SI's fidelity, we carried out preliminary experiments with correlation-preserving variance calculation via UP and approximating the truncation at branches.
However, we found that the error incurred by restricting the state representation to a small maximum number of paths (cf.~Assumption~\ref{enum:si_assumption_4}) dwarfs the benefits of these enhancements.
Figure~\ref{fig:si_assumptions_test} showcases this on a simple synthetic program (cf.~Appendix~\ref{app:synthetic_example}) by comparing the combination of the original SI with AD, its combination with UP (cf.~\eqref{eq:uncertainty_propagation}), and an unbiased stochastic approximation of the exact convolution.
While UP improves the fidelity of the derivative to the convolution, restricting the number of tracked paths introduces significant jumps and deviations from the reference.
The erratic results are caused by the decision which paths to merge, which is not smooth with respect to the program inputs: for a small change in the input value, an entirely different set of merging decisions may be made.
As a consequence of this observation, and since the state restriction dominates SI's computational cost, we abstain from exploring the above enhancements further and instead focus on the Restrict~\cite{chaudhuri2010smooth} algorithm.

\subsection{State Restriction Strategies}
\label{sec:state_restriction_strategies}

Tracking the effects of every possible branch in a program execution quickly leads to a state explosion that renders the execution of non-trivial programs intractable.
Thus, SI employs an algorithm to merge branches, enabling the restriction of the state size to a user-defined limit $M$.
The restriction is achieved by identifying and subsequently merging two elements of the Gaussian mixture, such that the cost defined as the deviation from the original overall distribution is minimized, which involves calculations of new means and standard deviations.
As noted in~\cite{chaudhuri2010smooth}, this algorithm is optimal in the sense that it minimizes the deviation from the overall original mixture.
However, the algorithm is computationally intensive, requiring an iteration across the variables of all combinations of path states to determine a pair of paths to merge.
Further, merging two dissimilar paths can result in high-variance mixture elements and unreachable program states even according to a strict probabilistic semantics.
For example, refer to line 7 in Fig.~\ref{fig:the_figure}.
If the two states were merged, this would result in $Z\sim\mathcal{N}(0.5,0.5)$ assuming that $w_1\,{=}\,w_2\,{=}\,0.5$.
In other words, $Z$'s two possible crisp integer values are merged into a single Gaussian, which can severely affect subsequent operations.
Here, we explore three alternate heuristics with different tradeoffs in fidelity and computational cost.

In the Restrict algorithm, the cost of merging two variables is determined by the difference in the mixture element's moments and by the paths' weights.
To avoid merging dissimilar states, we can select paths to merge solely on the moments and ignore the path weights.
We refer to this strategy as Ignore Weights (IW).

On the other hand, by only considering the paths' weights, the expensive pair-wise comparisons among the paths' states can be avoided.
In this strategy, which we refer to as Weights Only (WO), the weight is used as a proxy for each variable's contribution to the gradient.
Assuming sufficiently similar mean values across paths, the weight is a good indicator of a mixture element's contribution to the final expected value \eqref{eq:discrete_smooth_gradient}.
Most importantly, since every variable on a given path shares the same weight, a merge decision is reduced to determining the pair of paths with the lowest weights.
A more radical strategy that aims for improved performance and low variance at the same time is to discard (Di) paths with the lowest weights entirely, avoiding the merging of variables.
The disadvantage of this strategy is that value and gradient information from discarded paths is lost entirely, whereas merging of path states retains the available gradient information across all paths, albeit in an aggregated form.

Although IW, WO, and Di are suboptimal in the theoretical sense, we will see in our evaluation (cf.~\secref{sec:evaluation}) that the decrease in overhead and/or variance obtained through these strategies can lead to faster optimization progress than the strategy proposed by Chaudhuri et.\ al., which we abbreviate as Ch.

\subsection{Monte Carlo Approach to Smooth Differentiation}
\label{sec:monte_carlo_approach_to_smooth_differentiation}

The assumption of Gaussian distributions and restricting the state to a small subset of possible paths (cf.\ assumption~\ref{enum:si_assumption_4}) cause the results of SI to deviate from the exact convolution of the program's output with a Gaussian kernel.
As a consequence, SI's estimations of the program's output and its gradient may be significantly biased.
In the following, we present an alternative approximation of the probabilistic program semantics based on Monte Carlo sampling and AD.

The key idea is to revisit the decomposition of the program's convolution with a Gaussian into a sum over the control flow paths.
From \eqref{eq:discrete_smooth_gradient} it follows that the partial derivative wrt.\ dimension $k$ of the mean $\textbf{x}$ for a program with $N$ control flow paths is given by:
\begin{equation}\label{eq:discrete_smooth_derivative}
\begin{split}
  \pdv{\prog(X)}{x_k} & = \sum_{p=1}^{N} \pdv{\expval{Y_p} w_p}{x_k} \\
                      & = \sum_{p=1}^{N} \pdv{\expval{Y_p}}{x_k} w_p + \sum_{p=1}^{N} \expval{Y_p} \pdv{w_p}{x_k}.
\end{split}
\end{equation}

\noindent For readability, we abbreviate the probability of taking the control flow path $p$ with the weight $w_p\,{\equiv}\,\mathbb{P}(\mathfrak{P}\,{=}\,p)$ and the expected value of the output distribution conditioned on $p$ as $\expval{Y_p}\,{\equiv}\,\expval{Y|\mathfrak{P}\,{=}\,p}$.

We now consider the approximation using Monte Carlo sampling, i.e., through repeated execution of the program on inputs drawn from the input distribution.
Note that the presence of discontinuities disallows the interchange of the differentiation and expectation operators in $\partial{\expval{Y_p}}/\partial{x_k}$.
To still be able to rely on pathwise derivatives, we estimate $\expval{\partial{Y_p}/\partial{x_k}}$ and compensate for the effects of discontinuities separately.
The resulting estimator is similar to existing work based on smoothed pertubation analysis~\cite{gong1987smoothed,arya2022automatic}.
Averaging across the pathwise derivatives of samples restricted to path $p$ using the indicator function $\mathbf{1}$ leads to the IPA estimator:
\begin{equation}
  {\expval{\pdv{Y_p}{x_k}}} = \lim \limits_{S \to \infty} \frac{1}{n_p}
 \sum\limits_{s=1}^{S}{\pdv{y_s}{x_{k,s}}}\mathbf{1}_{\mathfrak{P}(\mathbf{x}_s)=p},
\end{equation}

\noindent where $S$ is the number of samples, the $x_{k,s}$ are sampled from $X_k\,{\sim}\,\mathcal{N}(x_k,\sigma^2)$, $y_s$ and $\mathfrak{P}(\mathbf{x}_{s})$ are the output and chosen control flow path when running $\prog$ on the sample $\mathbf{x}_s$, and $n_p$ is the number of samples on path $p$.

Each sample takes exactly one of the $N$ paths, and the weight $w_p$ of a path $p$ is its probability of being taken, i.e., as $S$ tends to infinity, $n_p / S$ approaches $w_p$.
Hence, the pathwise derivatives are accounted for simply by:
\begin{equation}
\begin{split}
  \sum_{p=1}^{N} \expval{\pdv{Y_p}{x_k}} w_p & = \lim \limits_{S \to \infty} \sum_{p=1}^{N} \frac{1}{n_p} \sum\limits_{s=1}^{S}{\pdv{y_s}{x_{k,s}}} w_p \mathbf{1}_{\mathfrak{P}(\mathbf{x}_{s}) = p}\\
                                             & = \lim \limits_{S \to \infty} \frac{1}{S} \sum\limits_{s=1}^{S}{\pdv{y_s}{x_{k,s}}}.
\end{split}
\label{eq:dgo_pathwise_derivatives}
\end{equation}

The remaining difficulty lies in the term $\partial w_p/\partial x_k$ from \eqref{eq:discrete_smooth_derivative}, which depends on the distribution of the branch conditions and their sensitivities to the program inputs.  Given a branch statement of the form ``if ($C$ \texttt{<=} $d$)``, with $d$ a constant, we refer to the random variable $B \coloneqq C - d$ as the branch condition.
The branch condition is true, i.e., the branch is taken, with probability $\mathbb{P}(B \le 0)$, which is the cumulative distribution function of $B$ evaluated at 0.
In SI, this probability is estimated based on the parameters of the assumed Gaussian distribution of $B$, and its derivative can thus be determined transparently via AD.
To determine the derivative in the general case, where $B$ may depend arbitrarily on $\mathbf{x}$, let $g$ be the function that describes the dependence of $B$ on $\mathbf{x}$, i.e.: $g(\mathbf{x}) \coloneqq B$.
Using the chain rule, we have:
\begin{equation}\label{eq:weight_derivative_chain_rule}
  \pdv{F_{g(\mathbf{x})}}{x_k} = \pdv{F_{g(\mathbf{x})}}{g(\mathbf{x})} \pdv{g}{x_k}.
\end{equation}

\noindent Here, $\partial F_{g(\mathbf{x})} / \partial g(\mathbf{x})\,{=}\,f_{g(\mathbf{x})}\,{=}\,f_B$ is the PDF of the condition, which we can estimate based on the samples, e.g., via kernel density estimation.
The second term $\pdv{g}{x_k}$ is the derivative of the branch condition wrt. $x_k$.
In a sampling-based regime, the value of this term at $0$ can be approximated by averaging the exact AD derivatives for realizations in a neighborhood $[-\delta, \delta]$.
Using the approximation of the product from \eqref{eq:weight_derivative_chain_rule} and denoting the sampling-based estimate of the branch conditions' PDF as $\tilde{f}_B$, we arrive at the following Monte Carlo estimator for the derivative of the path weight of the ``true`` case at a branch encountered on path $p$:
\begin{equation}\label{eq:weight_derivative_estimation}
  \pdv{w_p}{x_k} \approx \frac{-\tilde{f}_B(0)}{S_c} \sum_{s=1}^{S} \pdv{g}{x_{k,s}}(\mathbf{x}_{s})\mathbf{1}_{\mathfrak{P}(\mathbf{x}_{s}) = p, |g(\mathbf{x}_{s})| < \delta} 
\end{equation}
\noindent and analogously with positive sign in the ``false`` case, with $c := (\mathfrak{P}(\mathbf{x}_{s}) = p, |g(\mathbf{x}_{s})| < \delta)$.
The product of this term with an estimate of the program output near the branch accounts for the effect of a jump across the branch.
By combining Eq.~\ref{eq:weight_derivative_estimation} and Eq.~\ref{eq:dgo_pathwise_derivatives}, we arrive at our gradient estimator, which we refer to as the DiscoGrad Gradient Oracle (DGO).

Since densitities are estimated, the approach is compatible with program-inherent randomness across samples, the presence of which may reduce or eliminate the need for smoothing by perturbation of the inputs.

\begin{figure*}[t!]
  \centering
  \hspace{0.4cm}
\parbox{0.51\pdfpagewidth}{
  \subfloat[Original program using the DiscoGrad API.]{
\begin{lstlisting}[frame=none,label=lst:dg_example]
const int num_inputs = 1;
#include "include/discograd.hpp"

adouble _DiscoGrad_f(DiscoGrad<num_inputs> &dg,
                     aparams &p) {
  // user code
  sdouble x({p[0], dg.get_variance()}), y;
  $\hspace{-0.11cm}\colorbox{codehighlight}{\textbf{if} (x < 0) y = 0; \textbf{else} y = 1;}$
  return y.expectation();
}

int main(int argc, char **argv) {
  DiscoGrad<num_inputs> dg(argc, argv);
  DiscoGradFunc<num_inputs> func(_DiscoGrad_f);
  dg.estimate(func);
  return 0;
}
\end{lstlisting}}
  }\hspace{-2.3cm}\Huge\myarrow\hspace{0.4cm}
\parbox{0.3\pdfpagewidth}{
 \subfloat[Smoothed branch for the DGSI backend.]{
\begin{lstlisting}[backgroundcolor=\color{codehighlight},firstnumber=7,label=lst:dg_example_si]
...
si_stack.prepare_branch(x < 0);   
{ si_stack.enter_if(); y = 0; } 
{ si_stack.enter_else(); y = 1; }
...

\end{lstlisting}}\\
    \subfloat[Smoothed branch for the DGO backend.]{
\begin{lstlisting}[backgroundcolor=\color{codehighlight},firstnumber=7,label=lst:dg_example_dgo]
... 
dg.prepare_branch(x - 0);
if (x < 0) y = 0; else y = 1;     $$
...
\end{lstlisting}}\vspace{0.25cm}
}
\vspace{0.3cm}\hrule
\caption{Example DiscoGrad program (a) and the smoothed versions of the contained branch for SI (b) and DGO (c).}
\end{figure*}

Each sample $\mathbf{x}_s$ may encounter several branches along each dimension $k$.
For sufficiently large $\delta$, each $x_{k,s}$ may thus appear in the summation of \eqref{eq:weight_derivative_estimation} for multiple branches.
This situation is not accounted for since the contribution of each individual sample only captures the output's derivative on its encountered control flow path without explicitly considering the (transitive) effects of taking alternative paths.
We treat this case heuristically by assigning an affected sample the path weight derivative wrt. the relevant dimension of the branch with the most equal distribution of samples between $[-\delta, 0]$ and $(0, \delta]$.

For the problems considered in our evaluation, we found that it sufficed to set $\delta$ to $\infty$, indicating that the benefit of collecting more samples per dimension and branch outweighs the bias in the pathwise derivative introduced by choosing a larger neighborhood.
Even so, deeply nested branches can lead to only small numbers of samples being observed at each branch.
This problem can be mitigated by translating nested branches to sequential branches, which was straight forward for the problems considered in Section~\ref{sec:evaluation}.

\subsection{DiscoGrad: Smooth Differentiation of C\raisebox{0.1ex}{++} Programs}
\label{sec:discograd}

DiscoGrad is a tool to translate programs written in a subset of \cpp\ to a smooth representation in order to execute them according to an approximate probabilistic semantics and to estimate the smooth programs' gradient.
The tool is comprised of two main parts: a set of header-based back-ends that implement AD, SI, and AD-guided Monte Carlo gradient estimation on one hand, and source-to-source transformations implemented via the LLVM compiler toolchain to generate estimator-specific code that makes use of the respective back-end.
To allow for a meaningful evaluation of execution times, the code was carefully profiled and optimized using standard techniques such as early returns, avoiding unnecessary copy operations, and minimizing dynamic memory allocation.

Fig.~\ref{lst:dg_example} depicts a basic DiscoGrad program that implements the Heaviside function.
In the main function, instances of \texttt{DiscoGrad} and \texttt{DiscoGradFunc} are created as interfaces to the chosen estimator's backend.
In this example, the user function \texttt{\_DiscoGrad\_f()} initializes smooth variables of type \texttt{sdouble} using an input mean value and variance, branches on the smooth variable \texttt{x}, and returns the resulting expectation of \texttt{y}.
The listed program is the input to the smoothing transformation, which is applied to any user function prepended with the string \texttt{\_DiscoGrad\_}.
After compilation, the smoothed program outputs the expectation returned by the user function along with its gradient.

At present, besides crisp code, which remains unmodified, DiscoGrad's smoothing transformation supports mathematical operations and assignments on any combination of crisp and smooth variables, conditional branching, loops, functions on smooth variables, references, and pointers to smooth variables as well as simple uses of containers.
Among the features currently not implemented are smooth versions of the ternary operator, switch statements, and global variables.
DiscoGrad's features and limitations are documented in our repository\footnote{\url{https://doi.org/10.5281/zenodo.10288017}}, where the full source code and the programs used in the evaluation in \secref{sec:optimization_problems} can be accessed.

\subsubsection{AD Implementation}
\label{sec:ad_implementation}

DiscoGrad includes an implementation of forward-mode AD based on operator overloading.
At first appearance, reverse-mode AD might seem preferable, since it covers the common case of differentiating programs mapping large numbers of inputs to a single output in a single reverse pass.
However, forward-mode AD provides two key benefits in our problem setting: firstly, its memory consumption is independent of the number of operations carried out by the program.
In contrast, reverse-mode AD maintains a tape in memory that grows linearly with the number of operations.
Furthermore, in combination with SI, the memory consumption for the tape multiplies with the number of tracked control flow paths.
Secondly, forward-mode AD allows us to determine a variable's derivatives with respect to the inputs at any time throughout a program's execution without further cost, in contrast to the reverse passes that would be needed with reverse-mode AD.
This allows us to efficiently determine the derivatives of branch conditions to the inputs in our implementation of the Monte Carlo estimator described in \secref{sec:monte_carlo_approach_to_smooth_differentiation}.

In our implementation, a variables' tangents (partial derivatives) with respect to the inputs are carried along as arrays, allowing for compiler vectorization of the tangent operations\footnote{We verified that in the code generated by Debian clang, version 11.0.1-2, the tangent operations were translated to AVX instructions.}.
To exploit the frequent case of variables carrying at most one non-zero tangent, full tangent arrays are allocated lazily only once required.
A pool of tangent arrays is maintained to avoid frequent explicit memory allocations and deallocations when variables are created or destroyed.
While naive forward-mode AD incurs a slowdown factor equivalent to the input dimension, we will see in \secref{sec:execution_time} that these simple implementation-level design decisions suffice to reduce the AD overhead to a more tolerable level.

\subsubsection{SI implementation}

Executing a program according to the probabilistic semantics of SI deviates from a regular ``crisp'' execution in two main regards: firstly, when encountering a branching statement, both the \texttt{then} and the \texttt{else} case is visited.
Doing so repeatedly generates an exponential number of \emph{path states} representing the variables' values resulting from different branch sequences, which is restricted to a configurable maximum number to constrain the memory consumption and execution time.
Secondly, the mathematical, logical, and comparison operations of the original program are widened to operate on all present path states.
On each path state, operations originally carried out on scalars are executed on Gaussian distributions represented by their first two moments.
DiscoGrad takes a similar approach to the original implementation of SI in the EULER tool~\cite{chaudhuri2012euler}, but allows for the use of smoothed variables with \cpp\ features such as containers and references, integrates SI with AD, and implements the state restriction strategies presented in \secref{sec:state_restriction_strategies}.

The key idea is for the source-to-source transformation to flatten the program's control flow across all branches so that all branch bodies are visited and to delegate the management of the variable states in the different control flow paths to our back-end library.
In the back-end, the program state is managed by an instance of type \texttt{SiStack}, which holds the path states that are active at the programs' scopes, with the state in the currently visited scope (e.g., the \texttt{then}-body of an \texttt{if-else} statement) at the top (cf.~Fig.~\ref{lst:dg_example_si}).
The type \texttt{sdouble} (smooth double) overloads the operations defined for \texttt{double}, carries them out for all active paths in the current scope, and substitutes originally crisp mathematical operations by operations on the moments of Gaussians.
Each path state's weight is a floating-point variable subject to differentiation via our AD implementation.
Similarly, the mean and optionally also the variance of each Gaussian value representing a mixture element are differentiable variables.

Given an \texttt{if-else} statement with a condition \texttt{l <= r} where at least one of \texttt{l}, \texttt{r} evaluates to type \texttt{sdouble}, DiscoGrad determines on each active path state $p$ the conditional probability $\mathbb{P}_p(l - r \le 0)$ of entering the branch.
The other inequality operators are handled analogously.
For each existing path state, two new path states representing the \texttt{then} and \texttt{else} cases are created with the new weights determined by multiplying the conditional probability and its complement with the original path's weight.
Paths with weights below a configurable threshold (set to $10^{-20}$ in the evaluation) are discarded due to their negligible impact on the program's output and to avoid arithmetic underflow.
Smooth loop constructs are supported by exiting a loop once no path state with sufficient probability enters another iteration.

At a branch, DiscoGrad first halves the number of active paths using one of the state restriction strategies Ch, IW, WO, and Di described in \secref{sec:state_restriction_strategies} to make it possible to generate new paths.
As we will see in our evaluation, the overhead incurred by this step is decisive for the overall execution time under SI.
For the computationally expensive Ch strategy, we cache previously computed merge costs among the path states as long as they remain unchanged and maintain a priority queue to efficiently select the next pair of states to be merged.

\subsubsection{DiscoGrad Oracle (DGO) Implementation}

Our Monte Carlo estimator DGO (cf.~\secref{sec:monte_carlo_approach_to_smooth_differentiation}) estimates the gradient of the smoothed program based on a series of runs on perturbed inputs.
Since the individual samples follow the control flow of the original crisp program, an implementation of DGO is vastly more lightweight compared to SI.
Here, the type \texttt{sdouble} maps to a single scalar floating-point number differentiable via our AD implementation, without any probabilistic semantics.

The source-to-source transformation simply prepends each \texttt{if-else} statement with a condition of the form \texttt{l <= r} by a call that passes $l - r$ to our back-end, and analogously for the remaining inequality operators (cf.~Fig.~\ref{lst:dg_example_dgo}).
For each sample, the condition values in a neighborhood $[-\delta, \delta]$ and their derivatives are gathered in order to estimate the weight derivative according to \secref{sec:monte_carlo_approach_to_smooth_differentiation}.
The mean of the conditions' gradients is computed on the fly via AD, as is the overall pathwise gradient of the sample.
Having collected the branch conditions, we estimate the probability density of the condition values at each branch using kernel density estimation with a Gaussian kernel.

Subsequently, we iterate over the branches encountered by each sample to assign the sample the path weight derivatives along its path.
Finally, the pathwise gradients and the path weight derivatives are combined according to the summation of Equation~\eqref{eq:discrete_smooth_derivative} to yield the DGO estimate of the gradient.

\section{Evaluation}
\label{sec:evaluation}

Bringing everything together, we perform an extensive empirical evaluation of the proposed gradient estimators with respect to their execution times and fidelity, comparing them to two other estimators.
We combine them with the Adam gradient descent procedure to solve optimization problems and also compare against global optimization techniques.

\begin{table*}[t!]
  \footnotesize
  \centering
  \begin{tabular}{rlp{0.35\linewidth}p{0.31\linewidth}}
    \textbf{Problem} & \textbf{Randomness} & \textbf{Parameters} & \textbf{Objective Function}\\
    \toprule
    \traffic[d] & Deterministic & $d^2$ traffic light offsets with $d\,{\in}\,\{2,5,10,20,40\}$ & Traffic flow \\ \midrule
    \ac & Stochastic & $82$ neural network weights & Loss determined by deviation from target temperature and energy cost \\ \midrule
    \hotel & Stochastic & $56$ products' booking limits  & Revenue \\ \midrule
    \epidemics & Stochastic & Recovery rate, initial infection probability, $100$ location-specific infection probabilities & Mean squared error wrt.\ reference over steps and locations \\ \bottomrule
  \end{tabular}
  \caption{Overview of the benchmark problems.}
  \label{tab:problems}
\end{table*}

In all experiments, we distinguish two types of replications.
A single run of a program at a given solution is a \emph{microreplication}.
For stochastic programs, several microreplications are carried out, averaging the partial derivatives across microreplications.
A \emph{macroreplication} is a replication of an entire experiment (optimization, parameter sweep) starting from an initial solution and spanning a series of microreplications.
When multiple macroreplications are executed, all estimators are configured with the same sequence of starting solutions across macroreplications.

All measurements of execution times and optimization progress were carried out on a machine equipped with a 16-core AMD Ryzen 9 7950X processor and 64 GiB RAM running Debian GNU/Linux 11, running at most 16 processes in parallel.

\subsection{Selected Non-Smooth Optimization Problems}
\label{sec:optimization_problems}

Based on well-known optimization problems from the literature and practical applications, we implemented four evaluation problems dominated by discontinuous control flow.
An overview of the problems is provided in Table~\ref{tab:problems}.

\subsubsection{Traffic Lights}

The \traffic[d] problem is a deterministic macrosimulation of a \emph{road network} using a simple form of the cell transmission model~\cite{daganzo1994cell} covering a two-dimensional grid of four-way intersections with dimensions $d\times d$ over $d$ time steps.
Vehicles are represented as \emph{populations} (vehicle counts) per lane.
At each time step, $d$ new \emph{vehicles} are created at the northern and/or western border of the grid, each with a general movement direction to the opposite border interrupted by rare predetermined turns.
The traffic flow at each intersection is organized by a signal that alternates between green and red phases of two time steps each for the horizontal and vertical lanes, allowing at most one vehicle per step and lane to advance to the next intersection.
The parameters of this problem are the $d^2$ traffic light \emph{phase offsets} for each intersection and the objective is to maximize the total \emph{number of intersections passed} by the vehicles throughout the simulation.
Here, discontinuities arise from the discrete switching of the traffic signals.

\subsubsection{Air Conditioning}

The second problem, taking inspiration from~\cite{yang2022safe} and later referred to as \ac, considers the \emph{optimal control} of an air conditioning unit by a \emph{neural network}.
An insulated room with a single window is simulated over $10$ time steps.
Over time, the \emph{temperature} of the room gradually approaches the \emph{outside temperature} according to the room's \emph{insulation}.
With a probability of $5\%$ per step, the window is opened, decreasing the insulation drastically.
The task of the AC is to keep the temperature of the room as close to a chosen \emph{target} as possible by deciding on its on/off state and the cooling power ($2$ neural network outputs), given the target, previous, and new temperature together with its previous action ($5$ neural network inputs).
Each time the AC activates, an \emph{energy penalty} is incurred.
For each simulation, the initial, target, and outside temperature, as well as the insulation are chosen randomly to force the network to generalize.
Considering the feedback from the previous' time steps inputs and outputs, this problem can be viewed as training a recurrent neural network.
The problem parameters are the $82$ weights of the one layer deep network and the objective is the minimization of the loss function defined as the sum of the average mean squared error over time and the energy penalties.
Discontinuities arise both from the on/off state of the AC and the discrete randomness of the window opening event.

\subsubsection{Hotel Booking}

The third problem is a revenue maximization problem from the SimOpt benchmark suite \cite{simoptgithub,eckman2023simopt} and considers the \emph{optimal booking} of a hotel.
The hotel offers 100 rooms via 56 ``products'' reflecting the guests' arrival days within a week, the lengths of their stays, and two different rates: a rack rate and a discount rate.
Over the course of one week, guests arrive and request products according to per-product Poisson processes with predefined rates.
The number of bookings is restricted by a per-product \emph{booking limit}.
Whenever a requested product's booking limit is greater than zero, the guest can be accommodated, and the booking limits for all products are decremented to account for the reduced room availability on the days covered by the product.
Here, the goal is to maximize the revenue by adjusting the products' interdependent booking limits to the guests' arrivals.
The parameters are the 56 booking limits, each with an upper bound of 100.
Discontinuities are caused by the discrete decisions of whether a guest's request for a product can be satisfied.

\subsubsection{Disease Spread}

The final problem is an \emph{agent-based} SIR (susceptible, infected, recovered) simulation of \emph{disease spread} similar to~\cite{andelfinger2022towards}.
Over $25$ time steps, a population of $200$ individuals moves on an undirected \emph{graph topology} generated as a random geometric graph with $100$ nodes (locations) and average degree $\approx7$.
Initially, agents are infected with a certain probability.
Upon infection, a recovery time is scheduled with a delay drawn from an exponential distribution.
Throughout the simulation, agents move along predetermined paths along the edges of the graph, being infected by their neighbors at the same location and recovering at their scheduled time.
The probability of being infected depends on a per-location coefficient.
The $102$ parameters of this problem are the initial infection probability, the mean recovery time relative to the simulation end time, and the location-specific infection probabilities.
The objective is to fit a previously generated progression of the epidemic by minimizing the mean squared error between the distribution of agent states at each location and the states recorded in the reference trajectory.
Discontinuities arise due to the discrete randomness in the infections, which occur via Bernoulli trials, as well as the discrete recovery event.

\subsection{Gradient Estimators}
\label{sec:gradient_estimators}

The objective of the evaluation is to determine the utility of our gradient estimators DGSI and DGO for solving the optimization problems defined in the previous section by gradient descent.
We chose the popular Adam optimizer due to its well-known applicability to noisy gradient estimates\cite{kingma2014adam}.
As points of reference, we employ the PGO~\cite{nesterov2017random} and REINFORCE~(RF)~\cite{williams1992simple} estimators.
Smooth gradients are calculated by sampling over a set of normally distributed random perturbations of the program parameters, making REINFORCE also applicable to deterministic problems such as \traffic\ (cf.~Appendix~\ref{app:reinforce_derivation} for a complete derivation).
More formally, we use the following reference estimators to obtain the smoothed gradient $\smgrad_{\mathbf{x}}$:
\begin{align}
  \label{eq:pgo_estimator}
  \smgrad_{\mathbf{x}}^{\mathtt{PGO}}\,\prog(\mathbf{x}) & = \lim\limits_{S\to\infty}\frac{1}{S}\sum_{s=1}^S \frac{\prog(\mathbf{x}+\sigma\mathbf{u}_s)-\prog(\mathbf{x})}{\sigma}\mathbf{u}_s \\
  \intertext{\noindent and}
  \label{eq:reinforce_estimator}
  \smgrad_{\mathbf{x}}^{\mathtt{RF}}\,\prog(\mathbf{x}) & = \lim\limits_{S\to\infty}\frac{1}{S}\sum_{s=1}^S \frac{\prog(\mathbf{x}+\sigma\mathbf{u}_s)}{\sigma}\mathbf{u}_s,  
\end{align}

\noindent where $\mathbf{u}_s\,{\sim}\,\mathcal{N}(\mathbf{0}, \mathbf{I})$ are iid.\ variates of the standard multivariate normal distribution and $\sigma$ is the smoothing factor (i.e., standard deviation).
Notice how this formulation of REINFORCE, where the stochasticity is introduced by random perturbations, is very similar to PGO.

We note that, for stochastic problems, REINFORCE may also directly exploit the problem-specific stochasticity, but only if the log-derivative is known.
As our problems allow arbitrary probability distributions, the log-derivative is difficult to obtain in general.
Further, our proposed mechanisms and PGO work fully automatically, i.e., in a black-box setting.
We thus evaluate REINFORCE in the same setting.
Additionally, PGO can be combined with a random search as described in \cite{polyak1987introduction, nesterov2017random}.
The random gradient-free search algorithm described therein can be viewed as only taking one sample from PGO per step of descent.
In this evaluation, we limit our scope to performing \emph{gradient} descent, leaving the random search and possible interesting integrations with the Adam optimizer as future work (cf.~\secref{sec:conclusion}).

This leaves us with four optimization procedures. 
For brevity, we only show the estimator and number of samples (or tracked control flow paths), as they are all combined with Adam, for example ``PGO/100'' for PGO estimator with 100 samples or ``DGSI/Di/4'' for our SI implementation with four tracked control flow paths, using the Discard restriction strategy (cf.\ Table~\ref{tab:estimators}).

In the following, we perform three types of evaluation.
First, we evaluate the scalability in terms of computation time and memory (\secref{sec:execution_time}) and the gradient error (\secref{sec:gradient_fidelity}).
Testing the practical impacts of the former two, we conclude with an evaluation of the optimization performance (\secref{sec:optimization_performance}).
For the evaluation of optimization performance, we also compare two popular global optimization algorithms that would typically be applied to solve our non-smooth and non-convex problems: a standard \emph{genetic algorithm} (GA) with elitism (as provided by the \texttt{pyeasyga} module\footnote{\url{https://github.com/remiomosowon/pyeasyga}}) and \emph{simulated annealing} (SA) (by porting the version from the Ensmallen library\footnote{\url{https://ensmallen.org/docs.html\#simulated-annealing-sa}} to Python).
Our measurements of execution time and optimization progress over time exclude process startup times to avoid disadvantaging the existing baseline approaches without AD, which are typically faster and thus more strongly impacted by startup times.

\begin{table}[h]
  \begin{tabular}{rp{0.475\linewidth}p{0.244\linewidth}}
    \textbf{Name} & \textbf{Estimator} & \textbf{References}\\
    \toprule
    DGSI & DiscoGrad implementation of our combination of SI with AD using the & \secref{sec:approach} ff. and \cite{chaudhuri2010smooth} \\ 
    DGIS/Di & Discard restriction strategy (discard branches based on weight). & \\
    DGIS/WO & Weights Only restriction strategy (merge highest-weighted branches). & \\
    DGIS/IW & Ignore Weights restriction strategy (merge only based on moments). & \\
    \midrule
    DGO & DiscoGrad implementation of our Monte Carlo estimator. & \secref{sec:monte_carlo_approach_to_smooth_differentiation} \\ 
    \midrule
    PGO & Polyak's Gradient-Free Oracle. & \eqref{eq:pgo_estimator}; \cite{nesterov2017random}, \cite{polyak1987introduction} \\
    \midrule
    RF & DiscoGrad version of REINFORCE. & \eqref{eq:reinforce_estimator}; \cite{williams1992simple} \\ 
    \bottomrule
  \end{tabular}
  \caption{Overview of evaluated gradient estimators. All estimators are combined with the Adam optimizer.}
  \label{tab:estimators}
\end{table}

\subsection{Execution Time}
\label{sec:execution_time}

All of the AD-based gradient estimators incur an overhead over a crisp program execution without AD.
Here, we evaluate the scaling of the estimators' wall-clock execution time with the number of samples or paths.
Each measurement was repeated one hundred times, resulting in 95\% confidence intervals smaller than 7\% of any of the shown averages.

\begin{figure}[t!]
\includegraphics[width=0.99\columnwidth]{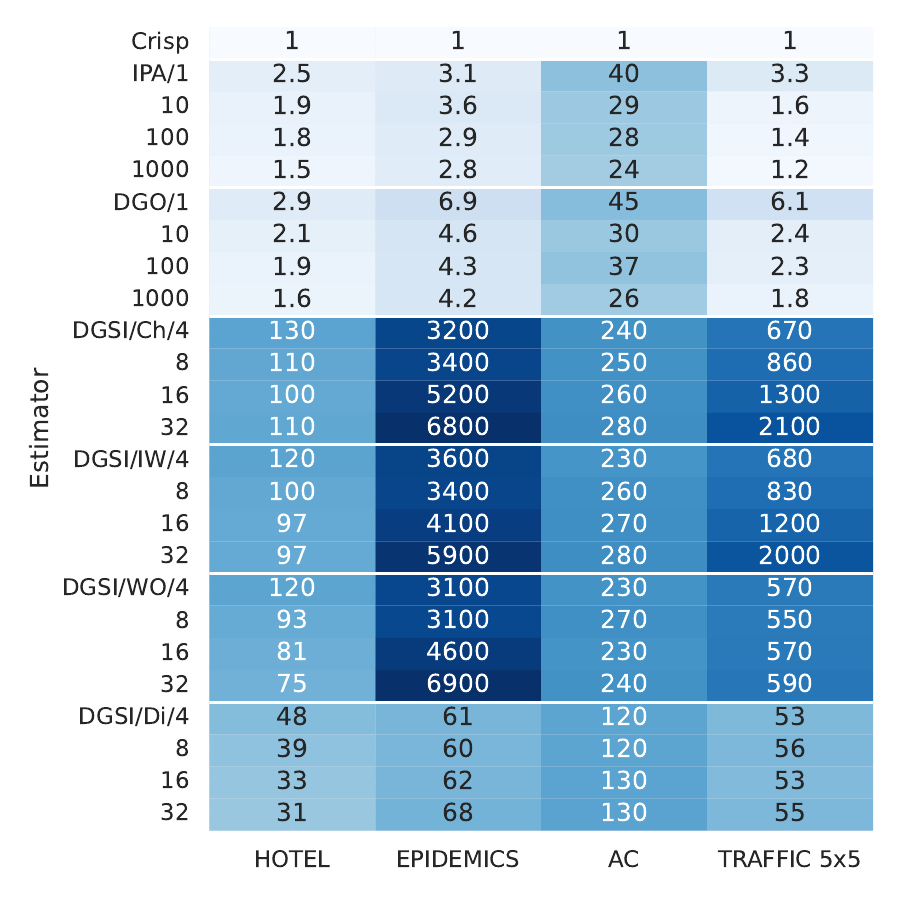}
\caption{Slowdown of the different gradient estimators compared to a crisp execution without AD, normalized to reflect the slowdown per sample (Monte Carlo estimators) or path (DGSI).}
\label{fig:slowdown}
\end{figure}

Figure~\ref{fig:slowdown} shows the slowdown of different configurations of the estimators over a single crisp program execution without AD, normalized to the number of paths or samples and rounded to two significant digits.
Each value can be interpreted as the slowdown per path or sample.

The overhead for the IPA estimator comprises only the cost of AD and the negligible cost of random number generation for the perturbations.
In the \epidemics, \hotel, and \traffic\ problem, we see the benefit of the simple sparsity optimization in our AD implementation.
In these problems, similar to the extreme case of the Heaviside function shown in Fig.~\ref{fig:ipa_fails}, the program's smoothed gradient depends mostly on the branches taken, with only limited arithmetic on variables that directly depend on the input parameters.
As a consequence, most variables do not carry a tangent value and the slowdown factor remains far below the input dimension, which would be the expected slowdown with naive forward-mode AD.
When increasing the number of samples, the reuse of tangent vectors (cf.~\secref{sec:ad_implementation}) allows the overhead per sample to gradually diminish.
In contrast, in the \ac\ problem, the output has a non-zero pathwise gradient with respect to the neural network coefficients.
Hence, most intermediate variables carry tangents with respect to all inputs and the slowdown becomes more pronounced.
Still, due to vectorization of the AD operations, the slowdown for IPA with 1\,000 samples is only about a quarter of the input dimension of 82.

Our Monte Carlo estimator DGO involves a configurable number of AD-enabled program executions, additional bookkeeping at branches, and kernel density estimations.
We see that accordingly, the slowdown is somewhat larger than IPA's.
However, we observe sublinear scaling behavior when increasing the number of samples.
Since branch condition values are stored and operated on per branch, the impact of the base costs for the per-branch operations diminishes with larger numbers of samples.
As with IPA, the \ac\ problem, which involves more input-dependent arithmetic operations, entails higher overhead.

SI incurs the cost for AD and carrying along Gaussian distributions across several control flow paths, as well as for restricting the state according to the chosen strategy.
Depending on the program and state restriction strategy, the number of control flow paths can fluctuate and may not always saturate the configured upper bound.
Hence, we normalize to the \emph{effective} number of control flow paths throughout the program's execution, which we define as the average number of paths active when encountering a branch.
Since the state restriction is applied before branches and dominates SI's execution time, normalizing to the number of paths at that point captures the main cost of the different SI estimators.
As expected, the slowdown of SI is much larger compared to the other estimators.
The restriction strategies based on merging paths (Ch, IW, WO) are up to two orders of magnitude more expensive than ``discard'' (Di).
The cost of selecting the next paths to be merged is negligible for WO, while it is quadratic in the number of paths for Ch and IW.
For all three of Ch, IW, and WO, the cost for the subsequent merging of paths is linear in the number of variables.
The results for the \epidemics\ problem, which uses the largest number of variables of the considered problems, show the resulting enormous overhead of the merging-based strategies, with only modestly better scaling using WO.
The DGSI estimators based on merging incurred the lowest overhead in the \hotel\ and \ac\ problems, which is a consequence of their comparatively smaller number of variables and branches.

Overall, a substantial slowdown is observed with all of the smoothing estimators, ranging from a factor of about two to several orders of magnitude compared to a single crisp execution.
In particular, the overhead of DGSI with the restriction strategies based on merging paths is likely to put many real-world applications out of reach.
Whether each estimator's overhead can be justified depends on the fidelity of the calculated gradients and the resulting progress in parameter synthesis problems, which we evaluate in the next sections.

\subsection{Gradient Fidelity}
\label{sec:gradient_fidelity}

In this section, we provide empirical measurements of the estimated smoothed gradient fidelity in terms of the mean absolute error, which is defined in the dimension $k$ for the input vector $\mathbf{x}$ as:
\begin{equation}
  \label{eq:mse}
  MAE(g, k) = \frac{1}{S}\sum_{s=1}^S \left|\smgrad^g_k\,\prog(\mathbf{x}_s) - \smgrad_k\,\prog(\mathbf{x}_s)\right|,
\end{equation}

\noindent where $\mathbf{x}_1,\dots,\mathbf{x}_S$ is a sequence of sample points and $\smgrad^g_k$ indicates the $k$-th component of the smoothed gradient estimate of the estimator $g\,{\in}\,\{$DGSI, DGO, PGO, RF$\}$, i.e., the partial derivative ${\partial\prog(\mathbf{x}_s)}/{\partial x_{k,s}}$.
To retrieve the MAE in dimension $k$, we uniformly sample from these partial derivatives in a problem-specific range in dimension $k$ around an optimal value for $\mathbf{x}$, as determined by optimization.

\begin{figure}[t!]
\includegraphics[width=0.99\columnwidth]{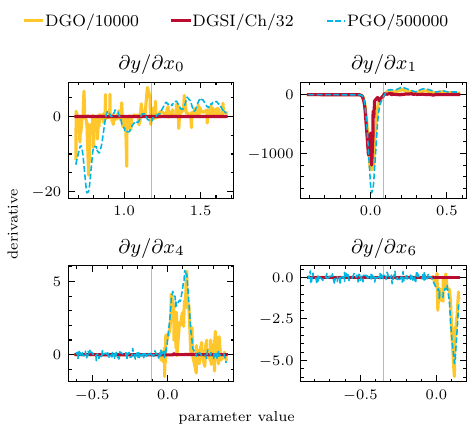}
\caption{Partial derivatives wrt. the recovery rate ($x_0$), initial infection probability ($x_1$) and location-specific infection probabilities $x_4$ and $x_6$ of the \epidemics\ model, as calculated by different gradient oracles. Using the PGO estimator with $500\,000$ samples as an unbiased baseline, the fidelity varies drastically among estimators, but also the problem dimensions. In particular, DGSI produces derivatives that are much too small.}
\label{fig:fidelity_epidemics}
\end{figure}

Evaluating this error is challenging, as for large problems it is expensive to calculate the exact smoothed gradient baseline $\smgrad\prog(\mathbf{x}_s)$.
Thus, we use a large number of samples ($5\times10^5$) of the unbiased PGO to produce a baseline with maximum $95\%$ confidence intervals of 0.003 (\ac), 0.014 (\traffic[2]), 0.033 (\traffic[5]), 0.425 (\hotel) and 7.3 (\epidemics). 
The wide maximum confidence intervals for the \hotel\ and \epidemics\ baselines can be attributed to some large partial derivatives in these problems (cf.~Fig.~\ref{fig:fidelity_epidemics}).
To reduce the computation time, we evaluate all problems in a deterministic setting by configuring a fixed seed and only consider the first $\leq25$ partial derivatives.

\begin{figure}[t!]
\includegraphics[width=0.99\columnwidth]{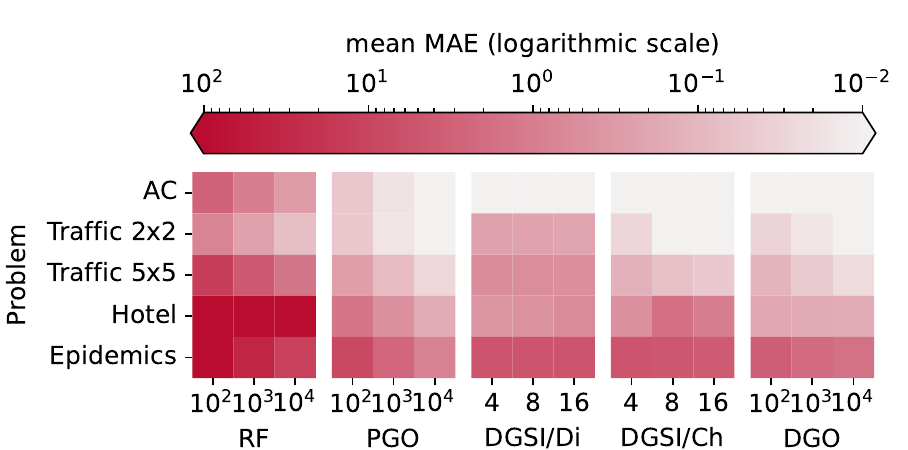} 
\caption{Error in the gradient estimate over different problems and estimator parametrizations (lower is better). Each cell reflects the mean absolute error (MAE) wrt. to the baseline (PGO/$500\,000$), averaged over the first $\leq25$ model dimensions.
The lowest error is provided by PGO and DGO, with the best result for REINFORCE magnitudes higher. For DGSI, a consistent decrease of the error with the number of paths can generally not be observed. In some cases, this is also true for DGO, signaling a potential bias in the estimate.\vspace{-0.4cm}}
\label{fig:fidelity_matrix}
\end{figure}

Fig.~\ref{fig:fidelity_matrix} shows an overview of the MAE wrt.\ the respective dimensions of the \ac, \traffic, \hotel\ and \epidemics\ problems.
The MAE, as indicated by the cell color intensity, is the average of the MAE defined in \eqref{eq:mse} over the first $k\,{=}\,1,\dots,25$ dimensions of each problem.
The first result is that, with some exceptions, the error of all sampling-based estimators decreases with the number of samples.
However, the estimators differ in how many samples they need to obtain the same fidelity, with our application of the REINFORCE estimator requiring orders of magnitude more samples than PGO and DGO.
Overall, our DGO estimator slightly outperforms the PGO estimation in terms of sample efficiency, although in some scenarios a bias prohibits further improvement with the number of samples.
The DGSI estimation is also very close to the baseline in many cases.
Especially on smaller problems such as \traffic[2], DGSI exhibits a competitive error, delivering good results even with only 8 tracked paths.
As expected, the Di restriction strategy is less accurate than the more expensive Ch.
Additionally, the error varies drastically between problems for all estimators.

\begin{figure}[t!]
\includegraphics[width=0.99\columnwidth]{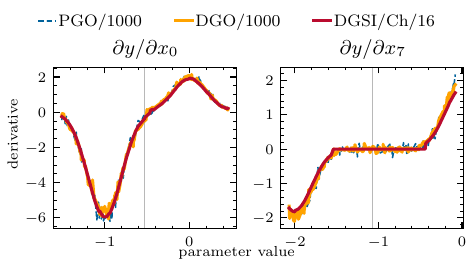}
\caption{Partial derivatives wrt.\ the signal offsets of the \traffic[5] problem, as calculated by different gradient estimates. Here and in the following figures, the centered vertical bar indicates the optimal value around which the samples were taken. In this fairly easy problem, all estimators deliver accurate partial derivatives.\vspace{-0.2cm}}
\label{fig:fidelity_traffic_5x5}
\end{figure}

\begin{figure}[t!]
\includegraphics[width=0.99\columnwidth]{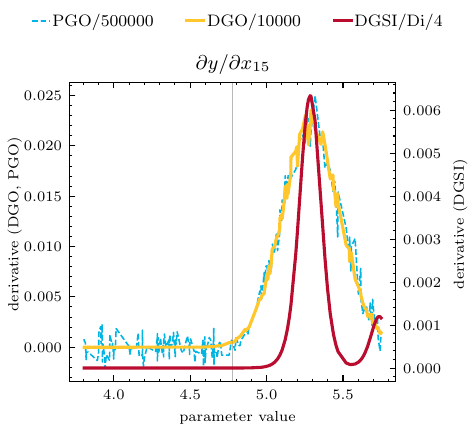}
\caption{Partial derivatives wrt.\ neural network parameters of the \ac\ problem, as calculated by different gradient oracles. In this problem, the DGO and DGSI estimators can profit from their accurate pathwise gradient, while PGO exhibits a lot of noise. The DGSI estimator is severely biased and thus plotted on a separate scale (right), but is able to capture the trends correctly.}
\label{fig:fidelity_thermostat}
\end{figure}

\begin{figure*}[b!]
\hspace{0.282cm}
\subfloat[Progress over time, y-axis zoomed for the last 10 min.]{\includegraphics[scale=0.9]{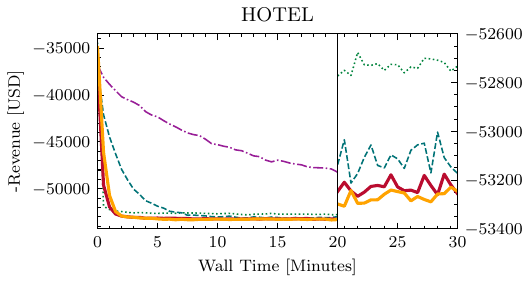}}
\hspace{0.38cm}
\subfloat[Progress over the first 500 optimization steps.]{\includegraphics[scale=0.9]{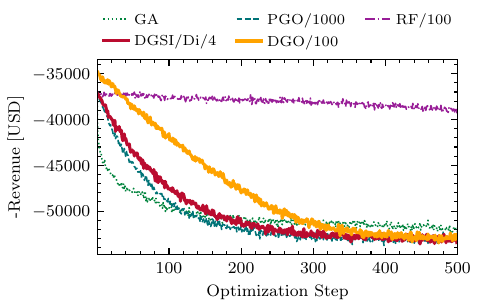}\label{fig:optimization_hotel_step}}
\caption{Optimization progress of the best-performing parametrization of each estimator for the \hotel\ problem over optimization steps.}
\label{fig:optimization_hotel}
\end{figure*}

A more granular view of these findings is depicted in Figures~\ref{fig:fidelity_traffic_5x5}, \ref{fig:fidelity_epidemics} and \ref{fig:fidelity_thermostat}, which show a comparison of selected partial derivatives as line plots.
The vertical bar in each plot represents the optimal value, around which the samples of the partial derivative were taken.
In Fig.~\ref{fig:fidelity_traffic_5x5} it can be seen that for relatively low-dimensional problems, DGO and DGSI deliver almost perfect results, and the optimum could be identified at the parameter values where the partial derivatives cross 0.
From Fig.~\ref{fig:fidelity_epidemics} it is evident that some problem dimensions pose greater challenges to smoothed derivative estimation than others.
For example, the estimates wrt.\ the location-specific infection probabilities $x_4$ and $x_6$ are much noisier than those wrt.\ the relative recovery time $x_1$; the estimates wrt.\ some of the dimensions seem to be biased for DGO and DGSI.
Additionally, DGSI delivers a gradient that is significantly smaller than the baseline and sometimes (erroneously) zero.
This can be attributed to the fixed size of the state tracked by SI, which necessarily results in a loss of smoothed derivative information for sufficiently large problems.
Interestingly, in the \ac\ problem, the gradients delivered by DGSI are also much smaller than the baseline, but can still capture the trend very well (cf.~Fig.~\ref{fig:fidelity_thermostat}).
On this problem, DGO is vastly less noisy than PGO, which can be attributed to the use of the exact pathwise derivative.

To conclude, we observe that with some exceptions where the gradient estimates are biased, DGO delivers accurate results.
Where it can exploit the pathwise derivative, the results exhibit less variance than the baseline PGO estimate.
DGSI is competitive in terms of the MAE on smaller problems with low dimensionality, beating every other estimate on \traffic[2], and obtains less noisy derivatives than the sampling-based estimators in these cases.
On larger problems, it incurs a bias, but can often still capture the underlying trend.
The effects of the observed differences in fidelity on the estimators' utility for gradient descent are evaluated in the next section.

\subsection{Optimization Performance}
\label{sec:optimization_performance}

\begin{figure*}[t!]
\hspace{0.8cm}
\subfloat[Progress over time, y-axis zoomed for the last 20 min.]{\includegraphics[scale=0.9]{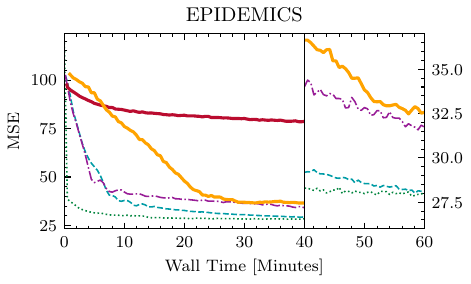}}
\hspace{1.275cm}
\subfloat[Progress over the first 50 optimization steps.]{\hspace{0.0cm}\includegraphics[scale=0.9]{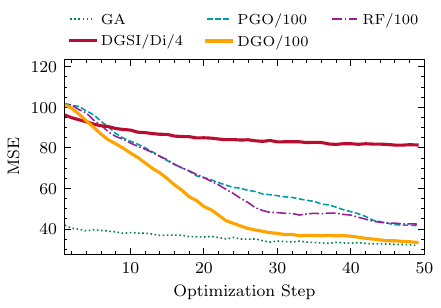}\label{fig:optimization_epidemics_step}}
\caption{Optimization progress of the best-performing parametrization of each estimator for the \epidemics\ problem.}
\label{fig:optimization_epidemics}
\vspace{-0.7cm}
\end{figure*}

\begin{figure*}[h!]
\hspace{0.85cm}
\subfloat[Progress over time, y-axis zoomed for the last 10 min.]{\includegraphics[scale=0.9]{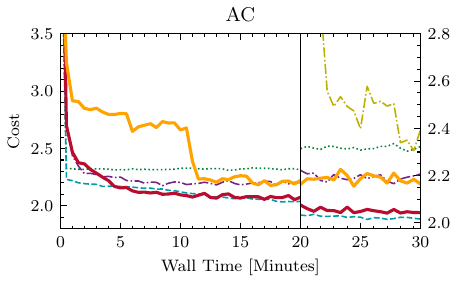}}
\hspace{1.66cm}
\subfloat[Progress over the first 800 optimization steps.]{\hspace{0.0cm}\includegraphics[scale=0.9]{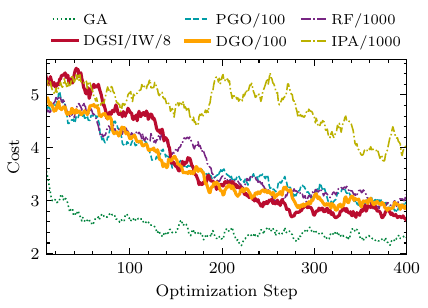}}
  \caption{Optimization progress of the best-performing parametrization of each estimator for the \ac\ problem.}
\label{fig:optimization_ac}
\vspace{-0.7cm}
\end{figure*}

\begin{figure*}[h!]
\hspace{0.525cm}
\subfloat[Progress over time, y-axis zoomed for the last 10 min.]{\includegraphics[scale=0.9]{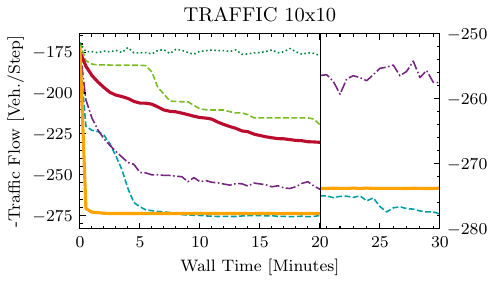}}
\hspace{0.95cm}
\subfloat[Progress over the first 500 optimization steps.]{\hspace{0.0cm}\includegraphics[scale=0.9]{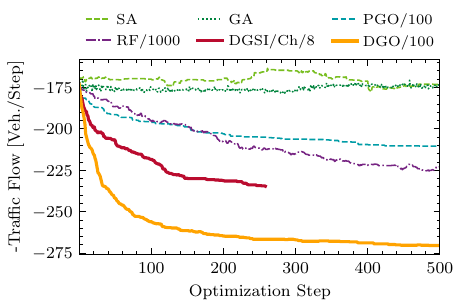}}
  \caption{Optimization progress of the best-performing parametrization of each estimator for the \traffic[10] problem.}
\label{fig:optimization_traffic_10x10}
\vspace{-0.7cm}
\end{figure*}

\begin{figure*}[h!]
\hspace{0.4cm}
\subfloat[Progress over time, y-axis zoomed for the last 40 min.]{\includegraphics[scale=0.9]{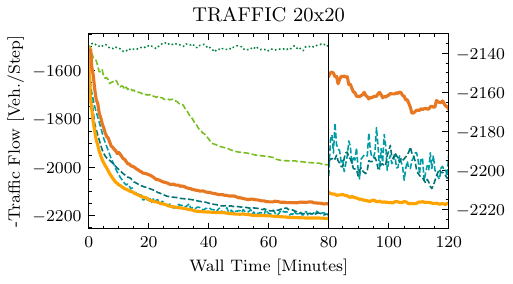}}
\hspace{0.66cm}
\subfloat[Progress over the first 200 optimization steps.]{\hspace{0.0cm}\includegraphics[scale=0.9]{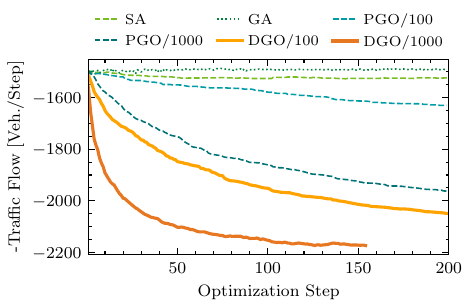}}
  \caption{Optimization progress of the best-performing parametrization of each estimator for the \traffic[20] problem.}
\label{fig:optimization_traffic_20x20}
\end{figure*}

\begin{figure*}[h!]
\hspace{0.26cm}
\subfloat[Progress over time, y-axis zoomed for the last 40 min.]{\includegraphics[scale=0.9]{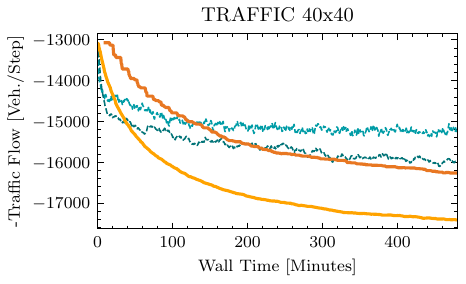}}
\hspace{1.38cm}
\subfloat[Progress over the first 50 optimization steps.]{\hspace{0.0cm}\includegraphics[scale=0.9]{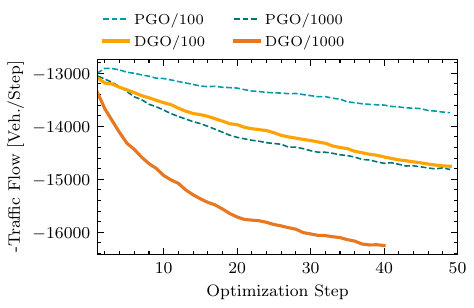}}
  \caption{Optimization progress of the DGO and PGO estimators for the \traffic[40] problem.}
\label{fig:optimization_traffic_40x40}
\end{figure*}

The optimization progress using the gradient-based approaches hinges on a suitable choice of the input standard deviation and the learning rate of the Adam optimizer.
Varying these two hyperparameters affects the degree to which the computed gradients accord with the original function on one hand and the ability to escape local minima on the other hand.
Here, we identified for each problem one combination $(\sigma_0, \eta_0)$ of standard deviation and learning rate where good progress was made with all estimators.
An automated hyperparameter sweep was then carried out covering three levels for each hyperparameter, covering all combinations of $\sigma_0 \cdot (\frac{1}{2}, 1, 2)$ and $\eta_0 \cdot (\frac{1}{2}, 1, 2)$.
As additional hyperparameters, we further varied the number of samples used by the stochastic gradient estimators, and for DGSI the number of paths and the path restriction strategy.
Across all problems and optimization methods, we carried out 3\,310 macroreplications, resulting in a total CPU time of about 3\,085 hours.

Where not stated otherwise, we show for each estimator the results of the hyperparameter combination that yielded the best solution at the end of the time budget.
At each solution determined via a smoothing estimator, we carried out an additional evaluation using the \emph{crisp} program.
The plots show the results of the crisp evaluation to ensure comparability of the solution qualities even if the smoothed program deviates from the crisp one.
After observing a premature convergence of simulated annealing (SA) to low-quality solutions, we decreased its hyperparameter $\lambda$, which determines the relative decrease in temperature per step, from its default value of $10^{-3}$ to $10^{-5}$.
Nevertheless, due to a lack of significant progress, the results for the SA are excluded from most plots.

Our plots show the optimization progress over wall-clock time and optimization steps.
Each data point in our results is the average of five macroreplications carried out for the respective combination of problem, estimator, and hyperparameters.
While the progress over time is the main concern for practical purposes, the progress over steps indicates the strides made when disregarding differences in execution time.
For the gradient-based estimators and SA, one step represents an evaluation of the objective function at one solution across the configured number of paths or samples for the smoothing estimators.
For the genetic algorithm (GA), one step represents an update from one generation to the next, which involves 50 function evaluations, one for each population member.

Figure~\ref{fig:optimization_hotel} shows the optimization progress for the \hotel\ problem.
Apart from SA and REINFORCE, all methods converge to a similar revenue of about 53\,200 within the time budget of thirty minutes.
The fastest convergence is achieved by the GA, albeit to a slightly worse solution than the best-performing methods.
Comparing the stochastic estimators, PGO performed best with 1\,000 samples, in contrast to 100 samples with DGO.
Figure~\ref{fig:optimization_hotel_step}, which shows the first 500 optimization steps, indicates that the DGO's higher variance with only 100 samples leads to less progress per step compared to PGO/1000.
DGSI performed best with the ``discard'' (Di) restriction strategy and with four paths, also converging to roughly the same solution quality as DGO and PGO.
Inspecting the solutions, we observed that all three of these methods arrived at similar final parameter combinations.

In the \epidemics\ problem (cf.~Figure~\ref{fig:optimization_epidemics}), PGO/100 and particularly the GA outperform the other methods.
We have seen in Section~\ref{sec:gradient_fidelity} that DGO and PGO both struggled to accurately estimate the gradient for this problem, in which there is a complex interplay between the initial infection probability, the recovery rate, and the per-location infection probabilities.
Here, GA converges extremely quickly, identifying a solution that is reached by PGO/100 only at the end of the time budget.
Considering the progress over the first 50 steps, we see that DGO makes larger strides than all other gradient estimators, indicating that its slower progress over time is a result of its higher execution time rather than lower-fidelity gradient estimates.
With DGSI, the convergence both over time and steps is too slow to be competitive.

Of our problems, \ac\ is the only one in which some of the partial derivatives are non-zero in the crisp case.
Hence, the classical IPA estimator can be applied, albeit without capturing the discontinuities generated by the decision whether cooling is activated in a time step.
As Figure~\ref{fig:optimization_ac} shows, all methods apart from SA were able to reduce the cost function to below 2.4, with the best solutions obtained by PGO/100 and DGSI with the IW strategy and eight paths.
IPA/1000 made very little initial progress, but approached the other methods' results at the end of the time budget.
Studying the \ac\ controller's behavior, all of the solutions obtained by the listed methods activate the cooling whenever the temperature is higher than the target.
However, the best two solutions identified by PGO and DGSI result in a more careful selection of the degree of cooling according to the current insulation and the energy cost.
Here, in contrast to the other problems, DGSI benefits from the existence of non-zero pathwise gradients, for which AD delivers exact values.

\begin{figure}[b!]
\includegraphics[width=0.99\columnwidth]{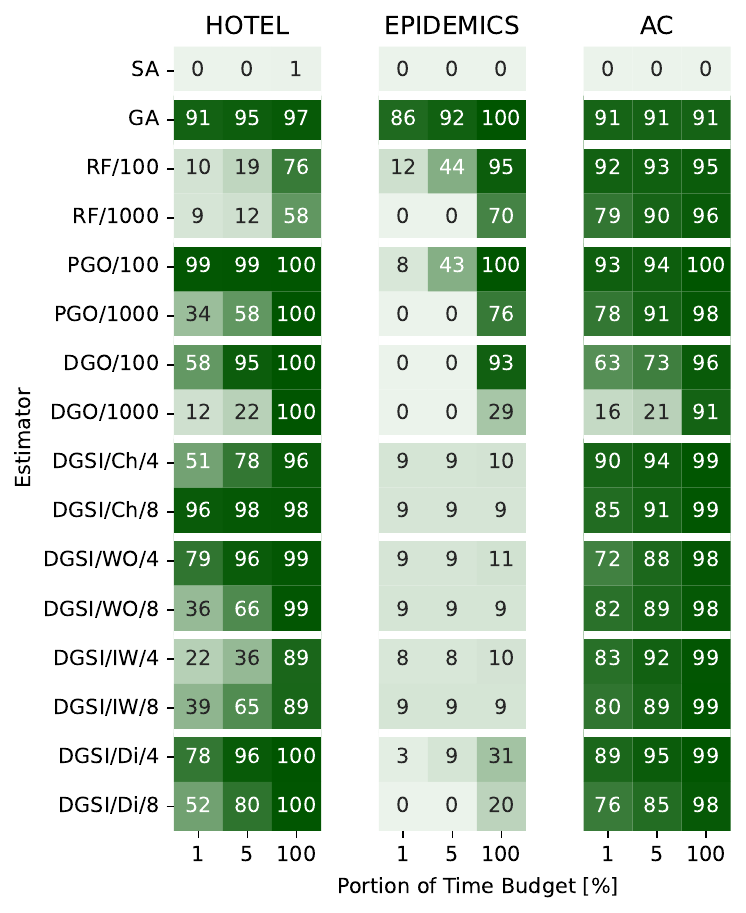}
\caption{Overview of the optimization progress over the time budgets for the \hotel, \ac\, and \epidemics\ problems.}
\label{fig:progress_heatmap_non_traffic}
\end{figure}

Finally, we consider the \traffic\ problem at the three scales of 10x10, 20x20, and 40x40 resulting in 100, 400, and 1\,600 decision variables.
Figure~\ref{fig:optimization_traffic_10x10} shows the results for the 10x10 grid.
In accordance with the fidelity results from Section~\ref{sec:gradient_fidelity}, where we have seen that DGO produces highly accurate gradients for this problem, the fastest convergence over the optimization steps by far is achieved by DGO/100, which still holds when plotted over optimization steps.
Here, GA was not able to improve beyond the initial solution.
In contrast to SA's results in the other problems, it is able to make substantial progress within the time budget, while still not being competitive with the best-performing methods.
REINFORCE is once again outperformed by PGO/100.
DGSI with the Ch strategy behaved somewhat similarly to SA, but having completed less than 300 steps was unable to obtain a competitive solution.

\begin{figure}[b!]
\includegraphics[width=0.99\columnwidth]{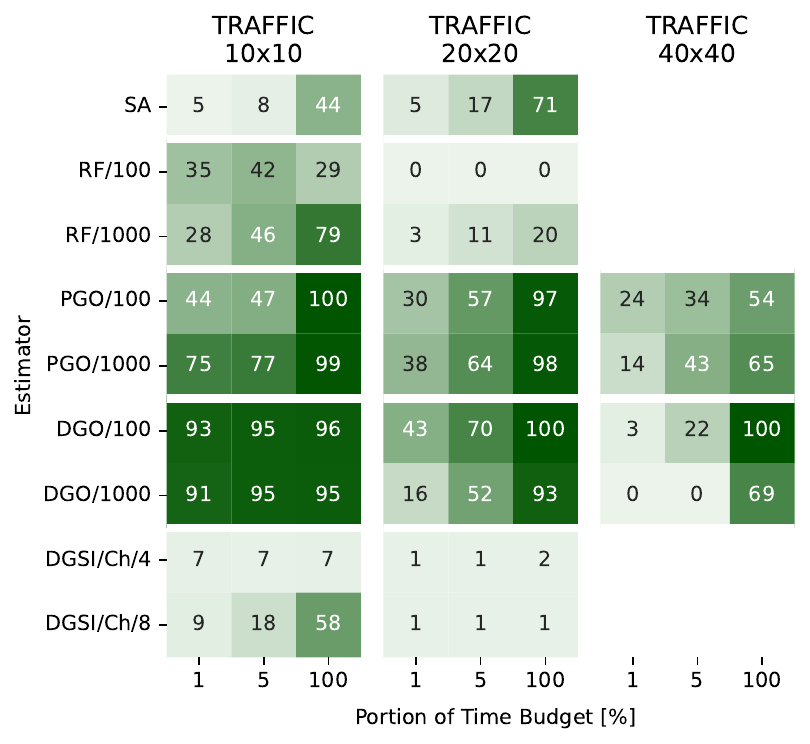}
\caption{Overview of the optimization progress over the time budgets for the \traffic\ problem.}
\label{fig:progress_heatmap_traffic}
\end{figure}

Similar trends are observed in Figure~\ref{fig:optimization_traffic_20x20} for our largest problem \traffic\ 20x20.
We omit DGSI and REINFORCE, which did not make substantial progress.
Here, DGO/100 provides the fastest convergence and the best solution, slightly better than PGO/100 and PGO/1000.
DGO/1000 exhibits vastly faster convergence over steps, but finished only about 150 steps within the time budget.

Since PGO and DGO consistently outperformed the other methods in the \traffic\ problem, we limited the computationally intensive experiment on the 40x40 grid to these two estimators with their respective best-performing hyperparameter combination from the 20x20 experiment.
Figure~\ref{fig:optimization_traffic_40x40} shows that for this problem with 1\,600 decision variables, DGO/100 vastly outperforms both PGO/100 and PGO/1000 over time, and DGO/1000 over optimization steps.
Here, DGO benefits from its use of AD to separate the effects of the individual input dimensions, whereas PGO must rely on the scalar program output alone.

The optimization progress measurements are summarized in the heatmaps shown in Figures~\ref{fig:progress_heatmap_non_traffic} and~\ref{fig:progress_heatmap_traffic}, which indicate each method's progress relative to the best improvement over the initial solution made by any method.
For the traffic problems, we only show the methods that made significant progress for at least one problem size.

In summary, DGSI has proven to swiftly determine high-quality solutions for the \hotel\ and \ac\ problems, in which the effects of choosing alternate branches are less extreme than in the other problems.
Particularly good results were seen for \ac\, which is the only of the considered problems in which the pathwise gradient is non-zero.
Generally, the stochastic estimators PGO and DGO delivered the most reliably high-quality solutions within the time budget.
While our AD-based estimator DGO showed outstanding performance, particularly in the \traffic\ problem with large numbers of input dimensions, the existing estimator PGO has the benefit of being applicable to existing programs without instrumentation.

\section{Conclusion}
\label{sec:conclusion}

Although an evaluation of gradient estimators targeting a problem domain as broad as parameter synthesis must necessarily be limited in scope, we identify several trends in our results.

The objective functions of the considered optimization problems are discontinuous and non-convex.
Nevertheless, local search based on gradient descent consistently outperformed global search via a genetic algorithm or simulated annealing.
Our results indicate that both for stochastic and non-stochastic problems, the local search is able to escape local minima sufficiently to swiftly identify high-quality solutions, likely due to the combination of noisy estimates with Adam.
Future work dedicated to more extensive benchmarking could consider more sophisticated global optimization methods such as recent trust region-based algorithms~\cite{shashaani2018astro} or metaheuristics~\cite{boussaid2013survey}.

Each of the studied gradient estimators comes with tradeoffs.
The estimators that combine smooth interpretation and automatic differentiation (DGSI) incur a substantial cost in execution time, depending on the state restriction strategy and the number of control flow paths carried along.
Notably, we saw that even if the fundamental approximations made by smooth interpretation were lifted, the need to combine or discard intermediate state severely impacts the gradient fidelity.
Accordingly, DGSI excels at problems with only limited branching or where the effects of branches are relatively constrained, i.e., non-chaotic problems.
Future efforts to improve the gradient estimation via smooth interpretation should thus focus on robust state restriction strategies.

Our proposed estimator using automatic differentiation and Monte Carlo sampling (DGO) is vastly less computationally expensive and avoids the various approximations made in smooth interpretation.
A key limitation is the need to obtain a sufficient number of samples at each branch, which may require many replications in the presence of deeply nested branches.
In the considered problems, where nested branching could be reformulated into sequential branching, the fidelity of DGO was always among the best of the considered estimators.
DGO's combination with Adam provided competitive convergence behavior for all problems, vastly outperforming its closest competitor in our highest-dimensional problem.
Our tool DiscoGrad offers an efficient implementation of DGSI and DGO to automate the estimations via DGSI and DGO for programs written in \cpp.

Considering the existing estimators, a remarkable observation is that Polyak's Gradient-Free Oracle (PGO), which does not require AD, exhibited low execution times and provided good results in all experiments.
We thus consider PGO and the closely related gradient-free algorithm Nesterov Random Search~\cite{nesterov2017random, larson2019derivative} promising alternatives to global search across high-dimensional parameter spaces, even for non-convex problems.

In our experiments, we carried out a hyperparameter sweep to identify suitable combinations of learning rates, degrees of smoothing, and numbers of samples.
Since these hyperparameters interact, our future work will include an exploration of scheduling algorithms that jointly select combinations of these hyperparameters and adjust them throughout the optimization process.

\section*{Acknowledgments}

We extend our thanks to Marian Zuska for implementing the \hotel\ optimization problem in DiscoGrad.
Funded by the Deutsche Forschungsgemeinschaft (DFG, German Research Foundation), grant no. 497901036.

\bibliographystyle{unsrt}
{
  \small
  \bibliography{references}
}

\newpage

\appendix

\section{REINFORCE for Discrete Programs}
\label{app:reinforce_derivation}

The following is a derivation of the REINFORCE estimator for deterministic programs $\prog$, which works by perturbing the input vector $\mathbf{x}$ with Gaussian noise.
Through the log-derivative trick, REINFORCE is defined as:
\begin{equation}\label{eq:reinforce_definition}
  \grad_\theta\expval[X\sim f_\theta(y)]{\prog(X)} = \expval[X\sim f_\theta(y)]{\prog(X)\grad_\theta\log f_\theta(X)},
\end{equation}

\noindent where $f$ is a density with parameter $\theta$.
In our case, we perturb the input vector $\mathbf{x}$ to $\prog$ with Gaussian noise to obtain a random variable $X\,{\sim}\,\mathcal{N}(\mathbf{x},\Sigma)$ where $diag(\Sigma)\,{=}\,\sigma^2$, so $\theta\,{\equiv}\,\mathbf{x}$ (cf.~\eqref{eq:smooth_program_as_convolution}).  Thus, we only need to derive the following gradient of $f_{\mathbf{x},\sigma}$, the normal density with mean $\mathbf{x}$ and variance $\sigma^2$:
\begin{equation}
\begin{split}
  \grad_{\mathbf{x}}\ln f_{\mathbf{x},\sigma}(y) & = \grad_{\mathbf{x}}\ln\left(\frac{1}{\sigma\sqrt{2\pi}}e^{-\frac{1}{2}\left(\frac{y-\mathbf{x}}{\sigma}\right)^2}\right) \\
                                                & = \grad_{\mathbf{x}}\ln\left(\frac{1}{\sigma\sqrt{2\pi}}\right) + \grad_{\mathbf{x}}\ln \left(e^{-\frac{1}{2}\left(\frac{y-\mathbf{x}}{\sigma}\right)^2}\right) \\
                                                & = \grad_{\mathbf{x}}\ln e^{-\frac{1}{2}\left(\frac{y-\mathbf{x}}{\sigma}\right)^2} \\
                                                & = -\grad_{\mathbf{x}}\frac{1}{2}\left(\frac{y-\mathbf{x}}{\sigma}\right)^{\!2} \\
                                                & = -\grad_{\mathbf{x}}\frac{(y-\mathbf{x})^2}{2\sigma^2} \\
                                                & = -\frac{2(y-\mathbf{x})}{2\sigma^2} \\
                                                & = \frac{y-\mathbf{x}}{\sigma^2}.
\end{split}
\end{equation}

\noindent Using this, we can approximate \eqref{eq:reinforce_definition} by Monte Carlo sampling:
\begin{equation}
  \expval[X\sim \mathcal{N}(\mathbf{x},\sigma^2)]{\prog(X)\grad_{\mathbf{x}}\log f_{\mathbf{x},\sigma}(X)} = \frac{1}{S}\sum_{s=1}^S \prog(\mathbf{x}_s)\frac{\mathbf{x}_s-\mathbf{x}}{\sigma^2},
\end{equation}

\noindent where $\mathbf{x}_s,\ s\,{\in}\,\{1,\dots,S\}$ are iid. variates of $X$.
To make the fact that $X$ is an ``artificial'' random variable obtained by perturbing the input vector $\mathbf{x}$ more explicit, it is convenient to reparametrize and redefine $X\coloneqq\mathbf{x}+\sigma U$ for $U\,{\sim}\,\mathcal{N}(\mathbf{0},\mathbf{1})$.
Applying this substitution to the previous equation yields:
\begin{equation}
  \begin{split}
    \grad_{\mathbf{x}}\expval[X]{\prog(X)} & = \frac{1}{S}\sum_{s=1}^S \prog(\mathbf{x} + \sigma \mathbf{u}_s)\frac{\mathbf{x} + \sigma \mathbf{u}_s-\mathbf{x}}{\sigma^2} \\
                                                                    & = \frac{1}{S}\sum_{s=1}^S \prog(\mathbf{x} + \sigma \mathbf{u}_s)\frac{\mathbf{u}_s}{\sigma},
  \end{split}
\end{equation}

\noindent where $u_s,\ s\,{\in}\,\{1,\dots,S\}$ are iid. variates of $U$.
This leads to the REINFORCE estimator shown in \eqref{eq:reinforce_estimator}.

\section{Implicit Covariance through Automatic Differentiation}
\label{app:implicit_covariance_through_ad}

In this appendix, we show how AD wrt.\ the program inputs can be used to implicitly account for the covariance when applying the standard uncertainty propagation formula.
This technique is also applied by the uncertainties-cpp library\footnote{\url{https://www.giacomopetrillo.com/software/uncertainties-cpp/doc/html/classuncertainties_1_1_u_real.html}}.
For clarity, we assume the common case of a binary function on two intermediate variables $a_1$ and $a_2$ that depend on each other via a single input variable $x$.
The derivation can easily be adapted to cases with multiple inputs and n-ary operations.
For the former, one needs to include the covariance between the components of the input vector (which in our case is always 0) in \eqref{eq:proposal} below, for the latter, one needs to consider $h(a_1,\dots,a_n)$.

Let the two intermediate variables $a_1\,{=}\,f(x)$ and $a_2\,{=}\,g(x)$ be linear functions $f$ and $g$ of the input variable $x$. 
The variances of $a_2$ and $a_2$ are computed from the input variance $\sigma_x$ according to \cite{benke2018error} to first order as
\begin{align}
  \label{eq:sigma_f_x}
  \sigma^2_{f\!(x)} & = \left(\frac{d f(x)}{d x}\right)^{\!2}\sigma^2_{x} \\
  \intertext{\noindent and}
  \sigma^2_{g(x)}   & = \left(\frac{d g(x)}{d x}\right)^{\!2}\sigma^2_{x}.
\end{align}

\noindent Because of the linearity of $f$ and $g$, this is exact.
If we apply a function $h$ on $a_1$ and $a_2$ and want to calculate the resulting uncertainty in the form of the variance, we need to account for their covariance \cite{benke2018error}:
\begin{equation}
\label{eq:to_show}
  \begin{split}
    \sigma^2_{h(a_1,a_2)} & \approx\left(\pdv{h(a_1,a_2)}{a_1}\right)^{\!2}\sigma^2_{a_1} + \left(\pdv{h(a_1,a_2)}{a_2}\right)^{\!2}\sigma^2_{a_2} \\
                          & + 2\left(\pdv{h(a_1,a_2)}{a_1}\right)\left(\pdv{h(a_1,a_2)}{a_2}\right)\mathrm{Cov}(a_1,a_2)
  \end{split}
\end{equation}

\noindent However, instead of tracking the covariance of each variable explicitly, the following equation can be used, which makes use of the derivative wrt.\ the input variable $x$:
\begin{equation}
  \label{eq:proposal}
  \sigma^2_{h(a_1,a_2)}\approx\left(\frac{d h(a_1,a_2)}{d x}\right)^{\!2}\sigma^2_x
\end{equation}

\noindent Under the usual assumption of linearity in $h$, this expression is equivalent to \eqref{eq:to_show}.
One can see this in our binary case by applying the chain rule, as used in AD, and expanding using the binomial theorem:
\begin{equation*}
  \begin{split}
    \sigma^2_{h(a_1,a_2)} & \approx \left(\pdv{h(a_1,a_2)}{a_1}\frac{d a_1}{d x} + \pdv{h(a_1,a_2)}{a_2}\frac{d a_2}{d x}\right)^{\!2}\sigma^2_x \\
                         & = \left(\pdv{h(a_1,a_2)}{a_1}\frac{d a_1}{d x}\right)^{\!2}\sigma^2_x \\
                         & + \left(\pdv{h(a_1,a_2)}{a_2}\frac{d a_2}{d x}\right)^{\!2}\sigma^2_x \\
                         & + 2\left(\pdv{h(a_1,a_2)}{a_1}\frac{d a_1}{d x}\right)\left(\pdv{h(a_1,a_2)}{a_2}\frac{d a_2}{d x}\right)\sigma^2_x
  \end{split}
\end{equation*}
We observe that, given the definition of $\sigma^2_{f(x)}$ in \eqref{eq:sigma_f_x}, we can substitute $(d a_1/d x)^2\sigma^2_x$ with $\sigma^2_{a_1}$ (analogously for $a_2$):
\begin{equation*}
  \begin{split}
    = & \left(\pdv{h(a_1,a_2)}{a_1}\right)^{\!2}\sigma^2_{a_1} + \left(\pdv{h(a_1,a_2)}{a_2}\right)^{\!2}\sigma^2_{a_2} \\
    + & 2\left(\pdv{h(a_1,a_2)}{a_1}\pdv{a_1}{x}\right)\left(\pdv{h(a_1,a_2)}{a_2}\pdv{a_2}{x}\right)\sigma^2_x.
  \end{split}
\end{equation*}

\noindent The last term of the sum can be brought into the form of \autoref{eq:to_show} by considering a first-order approximation of the covariance\footnote{\url{https://www.giacomopetrillo.com/software/uncertainties-cpp/doc/html/classuncertainties_1_1_u_real.html}} $\mathrm{Cov}(a_1,a_2)$ as 
\begin{equation}
\mathrm{Cov}(f(x),g(x)) \approx \frac{d f(x)}{d x}\sigma_x\frac{d g(x)}{d x}\sigma_x= \frac{d a_1}{d x}\frac{d a_2}{d x}\sigma^2_x
\end{equation}

\noindent Substitution then leads to \eqref{eq:to_show}.
Thus, assuming linear dependencies, AD can be used to avoid the need to explicitly track the covariance term.

\section{Synthetic Example}
\label{app:synthetic_example}

\begin{lstlisting}[caption={Program used to generate Fig.~\ref{fig:si_assumptions_test}.}]
const int num_inputs = 1;
#include "include/discograd.hpp"
double thresholds[]{
    0.270522,  0.897051,  -0.217456, -0.123758,
    0.644134,  0.839213,  -0.352062, -0.399846,
    0.998247,  0.933284,  0.910108,  -0.365462,
    -0.079554, 0.743998,  0.076774,  -0.790476,
    -0.715160, -0.835860, 0.642159,  0.311679,
    0.868213,  -0.435599, -0.222482, -0.085613,
    -0.873302, -0.916008, -0.059836, 0.879541,
    -0.168570, -0.054479, -0.674992, -0.253402
};

adouble _DiscoGrad_synthetic_test(
    DiscoGrad<num_inputs> &_discograd, aparams &p) {
  sdouble x({p[0], _discograd.get_variance()});
  sdouble y = x / 2.0; // correlate y with x
  sdouble v = x - y;   // here, y is correlated with x,
                       // necessitating DEA
  // this loop 'simulates' repeated branching behavior
  for (int i = 0; i < 32; ++i) {
    if (v < thresholds[i]) {
      // branches depend on previous branches
      v -= thresholds[i]; 
    }
  }
  return v.expectation();
}

int main(int argc, char **argv) {
  DiscoGrad<num_inputs> dg(argc, argv);
  DiscoGradFunc<num_inputs> func(
      _DiscoGrad_synthetic_test);
  dg.estimate(func);
  return 0;
}
\end{lstlisting}

\end{document}